\documentclass{article}

\usepackage[preprint]{neurips_2025}
\usepackage{multirow}
\usepackage{caption}
\usepackage{float}  % 在导言区加上这一行

\usepackage[utf8]{inputenc} % allow utf-8 input
\usepackage[T1]{fontenc}    % use 8-bit T1 fonts
\usepackage{hyperref}       % hyperlinks
\usepackage{url}            % simple URL typesetting
\usepackage{booktabs}       % professional-quality tables
\usepackage{amsfonts}       % blackboard math symbols
\usepackage{nicefrac}       % compact symbols for 1/2, etc.
\usepackage{microtype}      % microtypography
\usepackage{xcolor}         % colors
\usepackage{amsmath, amsfonts}
\usepackage{amsmath, amsthm}
\usepackage{algorithmic}
\usepackage{algorithm}
\usepackage{array}

\usepackage{color}
\usepackage{fancyhdr}
\usepackage{subfigure}
\usepackage{verbatim}
\usepackage{graphicx}
\usepackage{makecell}

\usepackage[normalem]{ulem}
\useunder{\uline}{\ul}{}

\usepackage{newfloat}
\usepackage{listings}
\usepackage{amssymb}

\usepackage{bbding}
\usepackage{pifont}
\usepackage{wasysym}
\usepackage{utfsym}

\newtheorem{definition}{Definition}
\newtheorem{theorem}{Theorem}
\newtheorem{lemma}{Lemma}

\title{Robust Non-negative Proximal Gradient Algorithm for Inverse Problems}

\author{%
  Hanzhang Wang\textsuperscript{1} Zonglin Liu\textsuperscript{1} Jingyi Xu\textsuperscript{2} Chenyang Wang\textsuperscript{3} Zhiwei Zhong\textsuperscript{4} Qiangqiang Shen\textsuperscript{4}\\
  \textsuperscript{1}Harbin Institute of Technology, Shenzhen\\
  \textsuperscript{2}Beihang University\\
  \textsuperscript{3}Harbin Institute of Technology\\
  \textsuperscript{4}City University of Hong Kong
}

\begin{document}

\maketitle

\begin{abstract}

Proximal gradient algorithms (PGA), while foundational for inverse problems like image reconstruction, often yield unstable convergence and suboptimal solutions by violating the critical non-negativity constraint. We identify the gradient descent step as the root cause of this issue, which introduces negative values and induces high sensitivity to hyperparameters. To overcome these limitations, we propose a novel multiplicative update proximal gradient algorithm (SSO-PGA) with convergence guarantees, which is designed for robustness in non-negative inverse problems. Our key innovation lies in superseding the gradient descent step with a learnable sigmoid-based operator, which inherently enforces non-negativity and boundedness by transforming traditional subtractive updates into multiplicative ones. This design, augmented by a sliding parameter for enhanced stability and convergence, not only improves robustness but also boosts expressive capacity and noise immunity. We further formulate a degradation model for multi-modal restoration and derive its SSO-PGA-based optimization algorithm, which is then unfolded into a deep network to marry the interpretability of optimization with the power of deep learning. Extensive numerical and real-world experiments demonstrate that our method significantly surpasses traditional PGA and other state-of-the-art algorithms, ensuring superior performance and stability.

\end{abstract}

\section{Introduction}
This paper focuses on the following convex optimization problems:
\begin{equation}
\min_x F(x), \quad \text{s.t. } x>0, \quad \text{where} \quad
\begin{cases}
F(x) = f(x), & \text{(Problem I)}, \\
F(x) = f(x) + g(x), & \text{(Problem II)}.
\end{cases}
\end{equation}

Here, $f$ is a convex and differentiable function, while $g$ is a convex but not necessarily smooth function. For Problem I (unconstrained convex and differentiable problem), researchers commonly use the classic gradient descent method for a solution \cite{ruder2016overview}. However, for Problem II (non-smooth composite optimization problem), which includes a non-differentiable term, researchers have explored various solution methods \cite{li2021decentralized}. The most common among these are splitting algorithms \cite{goldfarb2012fast}, which use first-order information to minimize the objective function. These include: the proximal gradient algorithm (PGA) \cite{li2015accelerated, salim2020wasserstein}, the alternating direction method of multipliers (ADMM) \cite{boyd2011distributed, hong2017linear}, the Douglas-Rachford splitting (DRS) \cite{eckstein1992douglas, patrinos2014douglas}, and the Pock-Chambolle (PC) algorithm \cite{chambolle2011first}. Among these, PGA is particularly popular due to its sound theoretical foundation and ease of optimization \cite{ dai2024proximal}.

The core idea of PGA is to perform a standard gradient descent step on $f$ followed by a proximal projection on $g$ \cite{laude2025anisotropic}. To accelerate convergence and enhance stability, researchers have introduced numerous improvements \cite{li2019decentralized,iutzeler2018proximal,si2024riemannian}. For instance, Keys et al. proposed the proximal distance algorithm, which combined classical penalty methods with distance majorization techniques \cite{keys2019proximal}. Additionally, Malitsky et al. introduced an adaptive proximal gradient method that leveraged the local curvature information of the smooth function to achieve full adaptivity \cite{malitsky2024adaptive}.

PGA provides a foundation for solving inverse problems in signal processing \cite{antonello2018proximal}, compressed sensing \cite{yao2025low}, and image reconstruction \cite{shen2011accelerated}. With the rise of deep learning, PGA has been successfully integrated into deep unfolding networks, creating a hybrid paradigm that integrates iterative optimization with learnable components to boost performance \cite{wei2022deep, mou2022deep}. This approach models the problem to be solved as an optimization objective and uses deep priors as the function $g$. In this area, Mardani et al. first proposed a novel neural proximal gradient descent algorithm that uses a recurrent ResNet to learn the proximal mapping, enabling high-resolution image recovery from limited sensory data \cite{mardani2018neural}. Xin et al. further improved deep unfolding networks by introducing an adaptive learning rate and borrowing the momentum technique from gradient descent, proposing a multi-stage and multi-level feature aggregation scheme for efficient MRI reconstruction \cite{xin2024rethinking}.

Despite the significant achievements of deep unfolding algorithms in vision tasks, their application still faces challenges. Their performance is often limited by the hyperparameter settings of the PGA, leading to unstable and suboptimal results. Furthermore, in vision tasks, images inherently have a non-negativity constraint. However, traditional PGA can produce negative solutions during the iterative process. Although these negative values may be numerically plausible, they violate the physical constraints of images and can exacerbate instability within the deep network during iteration.

\begin{figure}
\centering
 \includegraphics[width = 0.8\textwidth]{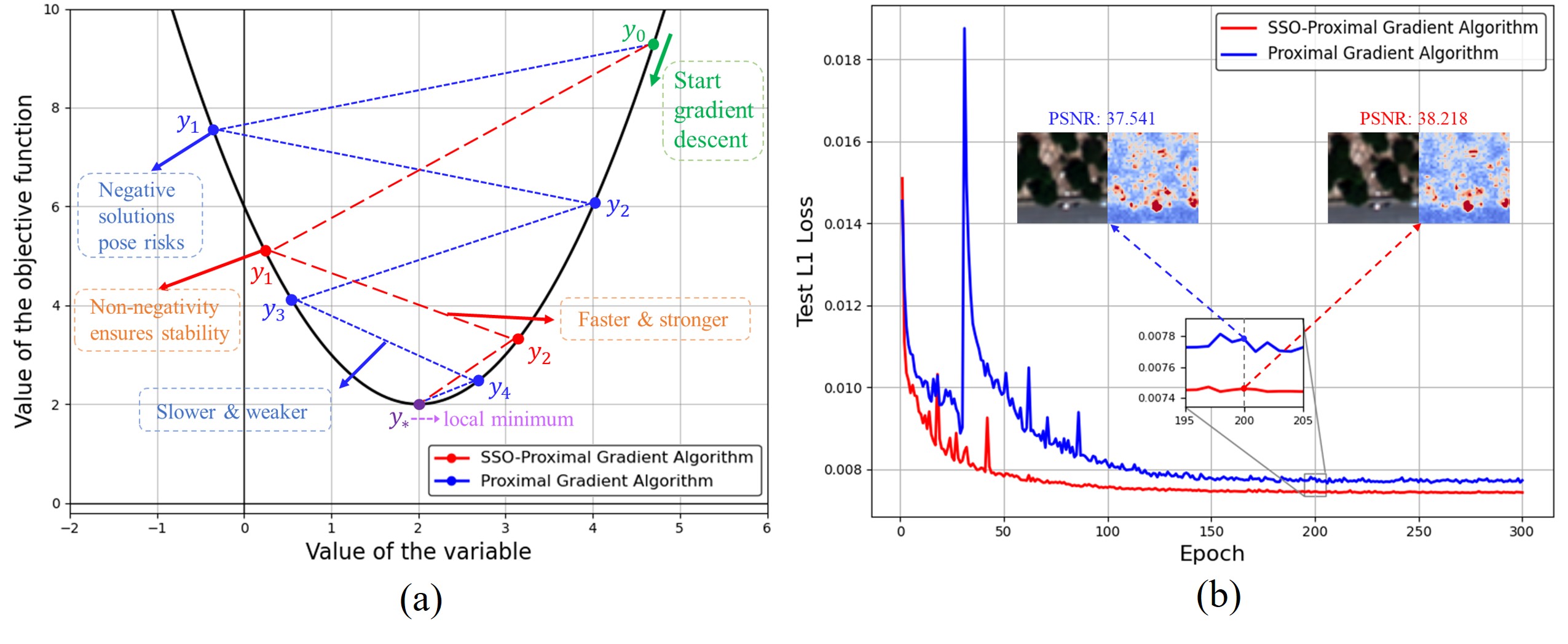}
\caption{Overview of our method. (a) A schematic comparison between PGA and SSO-PGA in the gradient descent process.
Compared to PGA, SSO-PGA benefits from the non-negativity constraint, yielding more stable solutions and demonstrating a faster convergence trajectory. (b) Comparison of test L1 loss curves between PGA and SSO-PGA on the WV3 dataset in the image fusion task over training epochs, with a zoomed-in view highlighting the reconstructed results at epoch 200.
SSO-PGA exhibits a more stable training process and achieves superior fusion quality.}
\vspace{-15pt}
\label{fig:motivation}
\end{figure}

To this end, we propose a novel robust non-negative proximal gradient algorithm (SSO-PGA), which maintains the optimization simplicity of the traditional PGA while effectively overcoming its drawbacks of instability and sensitivity. For Problem I, we reformulate the conventional additive gradient descent step into a new multiplicative update scheme via a Sliding Sigmoid Operator (SSO). Unlike traditional sensitive step sizes that often cause overshooting or vanishing updates, SSO adapts dynamically to the local gradient landscape, allowing finer control over the descent direction and magnitude. This leads to smoother convergence and mitigates abrupt changes. For Problem II, we can naturally extend the gradient descent algorithm from Problem I to the proximal gradient algorithm (SSO-PGA) by adding a proximal projection. Moreover, the inherent non-linearity and non-negativity of the SSO-PGA enhance robustness to noise. These properties make SSO-PGA particularly well-suited for vision tasks, as images inherently possess non-negative physical constraints. \textit{To our knowledge, this is the first work that improves upon PGA by using a multiplicative approach to fundamentally guarantee non-negativity and robustness, and adapt it to a deep network framework.} As shown in Fig.~\ref{fig:motivation}, compared with the existing PGA, SSO-PGA achieves more stable solutions, faster convergence, and superior performance, without introducing additional hyperparameters. The main contributions of this work are summarized as follows:
\begin{itemize}
  \item We propose a novel robust non-negative proximal gradient algorithm (SSO-PGA) with theoretical convergence guarantees,  which improves the gradient descent step of the traditional PGA via the Sliding Sigmoid Operator. This innovation inherently enforces non-negativity constraints, enhances nonlinear representation, and improves numerical stability. 
  \item Based on the proposed SSO-PGA, we develop a novel inverse problems model with efficient optimization. Specifically, we formulate Problem II as a multi-modal restoration problem and derive the corresponding optimization paradigm. This model is further unfolded into a structured deep neural network.
  \item Numerical experiments demonstrate superior performance for both Problem I and Problem II. Our deep unfolding network also shows a significant advantage in vision experiments, surpassing both the PGA baseline and other state-of-the-art (SOTA) algorithms for vision tasks. Moreover, compared to the PGA baseline, our SSO-PGA significantly improves convergence speed, hyperparameter stability, and robustness against perturbations.
\end{itemize}

\section{Related Work}
% \subsection{Proximal Gradient Algorithm}
Inverse problems are widespread across various fields, where one seeks to recover an unknown $\boldsymbol{y}\in \mathbb{R}^m$ from partial observations $\boldsymbol{x}\in \mathbb{R}^n$ \cite{deng2018variational,farahmand2024robust}. This is often based on a Gaussian noise assumption ($\boldsymbol{x} = \boldsymbol{H}\boldsymbol{y} +\boldsymbol{n}$) and can be represented as:
\begin{equation}
    \min_{\boldsymbol{y}} \|\boldsymbol{x}-\boldsymbol{H}\boldsymbol{y}\|_2^2,\quad(\text{Problem I}),
    \label{eq:inverse_problem_1}
\end{equation}
To achieve a more accurate recovery, researchers often introduce prior information \cite{nan2020deep}:
\begin{equation}
    \min_{\boldsymbol{y}} \|\boldsymbol{x}-\boldsymbol{H}\boldsymbol{y}\|_2^2 + \lambda f(\boldsymbol{y}),\quad(\text{Problem II}),
    \label{eq:inverse_problem}
\end{equation}
where $\boldsymbol{H}\in \mathbb{R}^{n\times m}$ is a degradation operator, and $f(\boldsymbol{y})$ is a regularization term that encodes prior knowledge about $\boldsymbol{y}$. When $f(\boldsymbol{y})$ is convex but possibly non-smooth (e.g., $\ell_1$-norm \cite{he2014new} or total variation \cite{palsson2013new}), the proximal gradient algorithm provides an efficient first-order method to solve the problem. Specifically, the update rule of the proximal gradient algorithm at the $t$-th iteration is given by \cite{beck2009fast}:
\begin{equation}
    \boldsymbol{y}^{t} = Prox_{f} \left( \boldsymbol{y}^{t-1} - \rho \nabla \mathcal{E}(\boldsymbol{y}^{t-1}) \right), \quad\mathcal{E}(\boldsymbol{y}^{t-1}) = \|\boldsymbol{x} - \boldsymbol{H}\boldsymbol{y}^{t-1}\|_2^2 ,
    \label{eq:PGA}
\end{equation}
where $\rho$ is a step size, and the proximal operator is defined as:
\begin{equation}
    Prox_{f}(\boldsymbol{v}) = \arg\min_{\boldsymbol{z}} \left\{ \frac{1}{2} \|\boldsymbol{z} - \boldsymbol{v}\|_2^2 + \lambda f(\boldsymbol{z}) \right\}.
\end{equation}
Although the proximal gradient algorithm enjoys fast convergence, it suffers from a major drawback: in imaging applications, pixel intensities are inherently non-negative, yet the update rule in Eq.~(\ref{eq:PGA}) may yield negative values. This not only violates the natural characteristics of images but also introduces vanishing gradient issues when implemented in deep unfolding networks. A straightforward solution to this problem is to restrict the update step by setting the step size $\rho$ as follows \cite{lee2000algorithms}:
\begin{equation}
\rho_i = \frac{\boldsymbol{y}^{t-1}_i}{(\boldsymbol{H}^{\top}\boldsymbol{H}\boldsymbol{y}^{t-1})_i}, \quad \text{for } i = 1, \dots, m.
\end{equation}

Substituting this into Eq.~(\ref{eq:PGA}) yields the following update rule:
\begin{equation} 
\begin{aligned} 
\boldsymbol{y}^{t}_i &= {Prox}_{f} \left( \boldsymbol{y}^{t-1}_i - \frac{\boldsymbol{y}^{t-1}_i}{(\boldsymbol{H}^{\top}\boldsymbol{H}\boldsymbol{y}^{t-1})_i} \left( (\boldsymbol{H}^{\top}\boldsymbol{H}\boldsymbol{y}^{t-1})_i - (\boldsymbol{H}^{\top}\boldsymbol{x})_i \right) \right)\ \\&= {Prox}_{f} \left( \frac{\boldsymbol{y}^{t-1}_i}{(\boldsymbol{H}^{\top}\boldsymbol{H}\boldsymbol{y}^{t-1})_i} (\boldsymbol{H}^{\top}\boldsymbol{x})_i \right) .
\end{aligned} \end{equation}

While this formulation guarantees non-negativity, it introduces a new numerical challenge: division by zero. Even when a small stabilization constant is introduced, this issue still results in numerical instability. This problem becomes more pronounced in deep unfolding networks, where it will prone to yield convergence failure or gradient explosion. 
% \subsection{Deep Unfolding Methods for Pan-sharpening}

\section{Method}
\subsection{SSO-enhanced Proximal Gradient Algorithm}
The motivation of this work is to address the non-negativity constraint in the proximal gradient algorithm while ensuring stability and robustness in both iterative optimization and deep learning frameworks. First, we give the definition of the Sliding Sigmoid Operator.

\begin{definition}[]
\label{df: SSO}
We define the Sliding Sigmoid Operator (SSO) as follows:
\begin{equation}
    SSO_{\alpha}(z) = 2\sigma(-z - \alpha) + 2\sigma(\alpha) - 1,
\end{equation}
where \( \sigma(c) = \frac{1}{1 + e^{-c}} \) denotes the sigmoid function, and \( \alpha \) is the  sliding parameter.
\end{definition}
\begin{minipage}[t]{0.48\textwidth}
\vspace{0pt}
As shown in Fig.~\ref{fig:sso}, SSO is essentially a sigmoid function augmented with a sliding parameter $\alpha$. Specifically, as $\alpha$ varies, the sigmoid curve slides along the coordinate point $(0, 1)$, adjusting its upper and lower bounds accordingly. Notably, the function always passes through the point $(0, 1)$, ensuring that its output is less than $1$ when the input is positive, and greater than $1$ when the input is negative. When the gradient is used as the input variable, this property, combined with the multiplicative update, naturally implements a gradient descent behavior. Furthermore, by adjusting $\alpha$, SSO adaptively controls the step size in the gradient descent process. Thereby, we can define the update rule of the SSO-enhanced proximal gradient algorithm in \textbf{Definition~\ref{df: 2}}:
\end{minipage}
\hfill
\begin{minipage}[t]{0.48\textwidth}
\vspace{0pt}
\centering
\includegraphics[width=\linewidth]{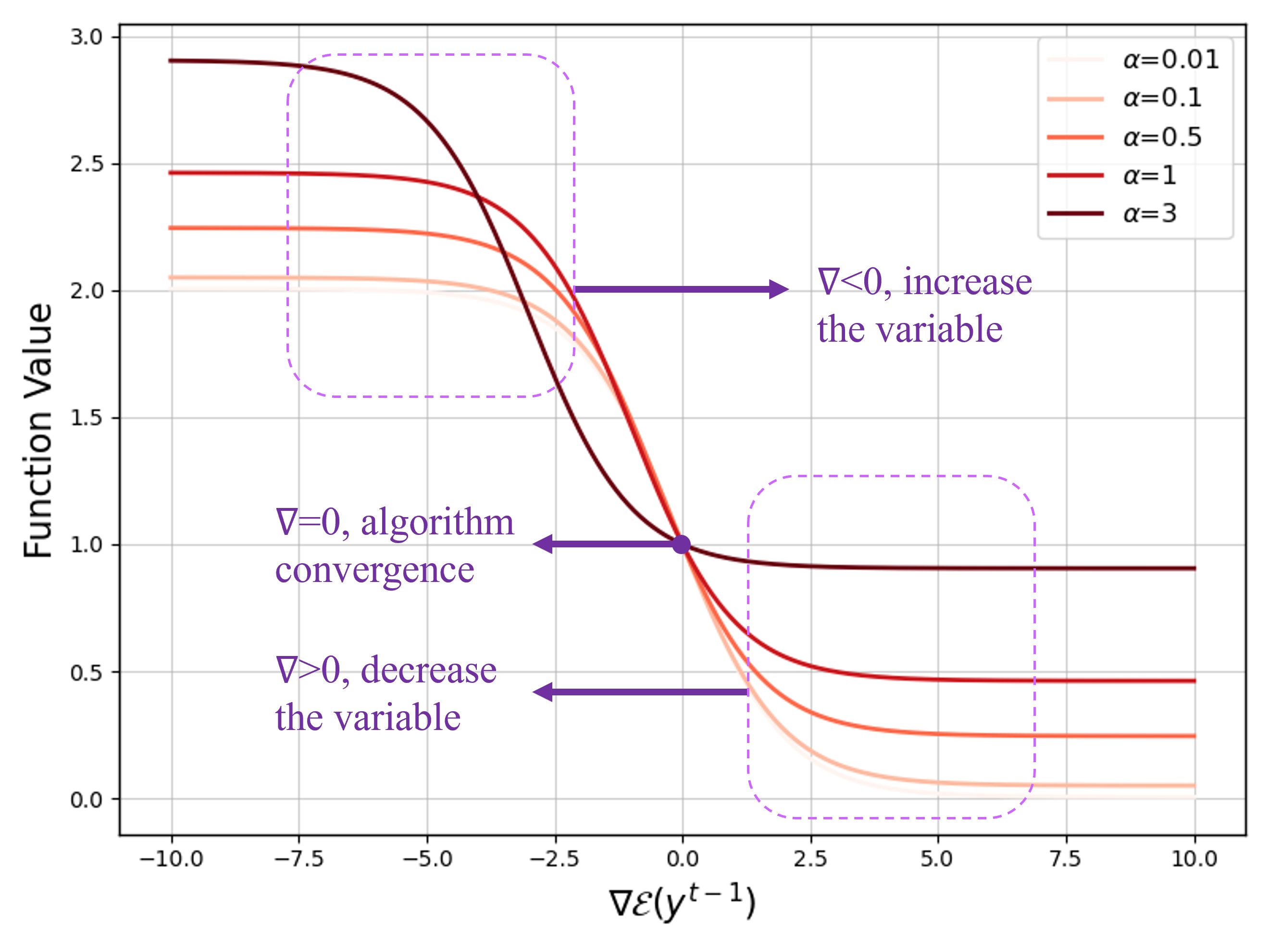}
\captionof{figure}{SSO working mechanism and its function curves under different $\alpha$ values.}
\label{fig:sso}
\end{minipage}

\begin{definition}[]
\label{df: 2}
The update rule of the SSO-enhanced proximal gradient algorithm (SSO-PGA) to the inverse problem in Eq.~(\ref{eq:inverse_problem}) at the t-th iteration is defined as follows:
\begin{equation}
\begin{aligned}
\boldsymbol{y}^t &= \boldsymbol{y}^{t-1}\odot SSO_{\alpha}(\nabla \mathcal{E}(\boldsymbol{y}^{t-1})),\quad &&(\text{For Problem I}),\\
    \boldsymbol{y}^t &= Prox_{f}\left(\boldsymbol{y}^{t-1}\odot SSO_{\alpha}(\nabla \mathcal{E}(\boldsymbol{y}^{t-1}))\right),\quad &&(\text{For Problem II}),
\label{eq: ssoupdate}
\end{aligned}
\end{equation}
\end{definition}
where $\odot$ denotes the element-wise product. Through SSO-PGA, we not only preserve the original gradient descent mechanism, but also constrain the updated variable within a multiplicative range of \( (2\sigma(\alpha) - 1,\, 2\sigma(\alpha) + 1) \) relative to the original variable, thereby enabling more robust gradient descent.
Moreover, the SSO multiplier enforces non-negativity of the updated variable, which better aligns with the characteristics of natural images. Then, we provide the following \textbf{Theorem~\ref{thm:1}}.
\begin{theorem}
\label{thm:1}
There exists \( \rho_i > 0 \) such that the SSO update rule is equivalent to a standard gradient descent step:
\begin{equation}
    \boldsymbol{y}_i^t = \boldsymbol{y}_i^{t-1} \cdot {SSO}_{\alpha}\left(\nabla \mathcal{E}(\boldsymbol{y}_i^{t-1})\right)= \boldsymbol{y}_i^{t-1} - \rho_i \nabla \mathcal{E}(\boldsymbol{y}_i^{t-1}),\quad \text{for } i = 1, \dots, m.
\end{equation}
\end{theorem}

Please refer to the \textit{APPENDIX} for the proof. \textbf{Theorem~\ref{thm:1}} demonstrates that SSO-PGA retains the fundamental logic of traditional gradient descent. SSO-PGA integrates the nonlinear representational capacity of the Sliding Sigmoid Operator with the theoretical foundation of gradient descent, enabling it to maintain stability while offering greater flexibility for adaptive adjustment.

Here, we prove the convergence of SSO-PGA. As the proximal step is unchanged from PGA, we only prove the gradient descent part. First, we introduce three lemmas.
\begin{lemma}
\label{lemma1}
For every \( \alpha\ge 0 \) and every \( z\in\mathbb R \), the following hold:
\begin{equation}
    \displaystyle\bigl|SSO_{\alpha}(z)-1\bigr|\le \eta(\alpha)\,|z| ,\quad\eta(\alpha)=\frac{1+\alpha}{2}.
\end{equation}
\end{lemma}

\begin{lemma}
\label{lemma3}
Let \(\mathcal E\colon\mathbb R^{n}\to\mathbb R\) have \(L\)--Lipschitz gradient.  Then for any \(\boldsymbol{y},\boldsymbol{d}\in\mathbb R^{n}\) \cite{nesterov2013introductory}:
\begin{equation}\label{eq:descent-lemma}
  \mathcal E(\boldsymbol{y}+\boldsymbol{d})\le\mathcal E(\boldsymbol{y})+\langle \nabla\mathcal E(\boldsymbol{y}),\boldsymbol{d}\rangle+\frac{L}{2}\,\|\boldsymbol{d}\|_2^{2}.
\end{equation}
\end{lemma}

\begin{lemma}
\label{lemma2}
Given $\mathcal E(\boldsymbol{y}) =\|\boldsymbol{x}-\boldsymbol{H}\boldsymbol{y}\|_2^2$, for all \(\boldsymbol{y},\boldsymbol{z}\in\mathbb R^{n}\), the following hold:
\begin{equation}
  \bigl\|\nabla\mathcal E(\boldsymbol{y})-\nabla\mathcal E(\boldsymbol{z})\bigr\|_2\le L\|\boldsymbol{y}-\boldsymbol{z}\|_2,\quad L=2\|\boldsymbol{H}\|_2^{2}.
\end{equation}
\end{lemma}

\begin{theorem}
\label{theorem2}
%Let $L=2\|\boldsymbol{H}\|_2^{2}$ and 
Let $0\le \alpha \le 2/(\kappa\|\boldsymbol{H}\|_2^{2})-1$, the inverse problem \(\|\boldsymbol{x}-\boldsymbol{H}\boldsymbol{y}\|_2^2\) is nonincreasing under the update rule:
\begin{equation}
    \boldsymbol{y}^t = \boldsymbol{y}^{t-1}\odot SSO_{\alpha}(\nabla \mathcal{E}(\boldsymbol{y}^{t-1})),
\label{eq: proof1}
\end{equation}
where $\kappa = \|\boldsymbol{y}^{t-1}\|_{\infty}$.

\end{theorem}
Please refer to the \textit{APPENDIX} for the proof. 
From \textbf{Theorem~\ref{theorem2}}, combined with the fact that \(\mathcal{E}(\boldsymbol{y}^{t}) \ge 0\) for every \(t \ge 1\), we can conclude that the inverse problem \(\|\boldsymbol{x} - \boldsymbol{H}\boldsymbol{y}\|_2^2\) converges to a local minimum under the gradient descent rule based on the SSO. It is worth noting that the condition \(0\le
\alpha \le 2/(\kappa\|\boldsymbol{H}\|_2^{2})-1,
\) used in the proof is merely a sufficient condition for ease of analysis. In experiments, we have found that \(\alpha\) admits a much broader range of values.

\subsection{Formulation and Optimization}
We formulate the SSO-PGA framework for solving inverse problems. Using multi-modal restoration as an example, given an observed image $\boldsymbol{\mathcal{X}}\in \mathbb{R}^{h\times w\times C_1}$ and a guided image 
$\boldsymbol{\mathcal{Y}}\in \mathbb{R}^{H\times W\times C_2}$, our goal is to reconstruct the target image $\boldsymbol{\mathcal{H}}\in \mathbb{R}^{H\times W\times C_1}$. We explicitly model the degradation processes in both domains to capture the differences between different modalities: 
\begin{equation}
\begin{aligned}
\min_{\boldsymbol{\mathcal{H}}, \boldsymbol{\mathcal{T}}}
\|\boldsymbol{\mathcal{X}} -\boldsymbol{\mathcal{K}}\boldsymbol{\mathcal{H}}\|_F^2 +\beta\| \boldsymbol{\mathcal{Y}} -\boldsymbol{\mathcal{S}}\boldsymbol{\mathcal{T}}\|_F^2,
\end{aligned}
\end{equation}
where $\boldsymbol{\mathcal{T}}\in \mathbb{R}^{H\times W\times C_1}$ denotes the guided-aligned latent embedding of the target image. $\boldsymbol{\mathcal{K}}$ and $\boldsymbol{\mathcal{S}}$ represent different degradation operators. We further enforce cross-domain consistency between the target image features and their guided-aligned embedding, thereby jointly preserving details in both domains:
\begin{equation}
\begin{aligned}
\min_{\boldsymbol{\mathcal{H}}, \boldsymbol{\mathcal{T}}}
\|\boldsymbol{\mathcal{X}} -\boldsymbol{\mathcal{K}}\boldsymbol{\mathcal{H}}\|_F^2 +\beta\| \boldsymbol{\mathcal{Y}} -\boldsymbol{\mathcal{S}}\boldsymbol{\mathcal{T}}\|_F^2+\gamma\|\boldsymbol{\mathcal{T}} - f(\boldsymbol{\mathcal{H}})\|_F^2,
\end{aligned}
\end{equation}
where $f(\cdot)$ is a feature transformation network. Finally, a deep prior $\phi(\cdot)$ is incorporated to further enhance the reconstruction quality of the target image. The final optimization objective can be formulated as:
\begin{equation}
\begin{aligned}
\min_{\boldsymbol{\mathcal{H}}, \boldsymbol{\mathcal{T}}}
\|\boldsymbol{\mathcal{X}} -\boldsymbol{\mathcal{K}}\boldsymbol{\mathcal{H}}\|_F^2 +\beta\| \boldsymbol{\mathcal{Y}} -\boldsymbol{\mathcal{S}}\boldsymbol{\mathcal{T}}\|_F^2+\gamma\|\boldsymbol{\mathcal{T}} - f(\boldsymbol{\mathcal{H}})\|_F^2+\phi({\boldsymbol{\mathcal{H}}}).
\end{aligned}
\label{eq: model}
\end{equation}
\begin{figure}
\centering
 \includegraphics[width = \textwidth]{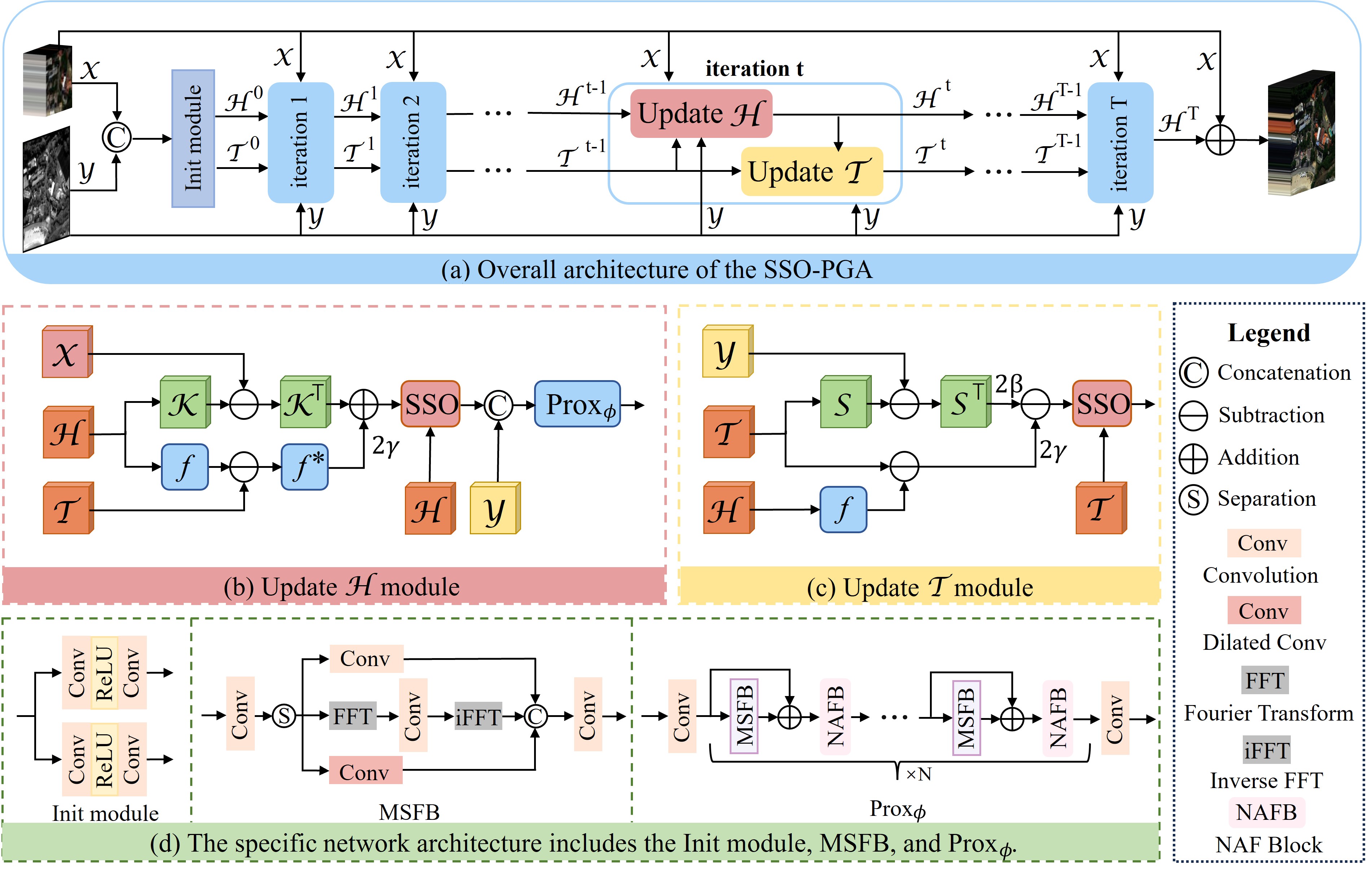}
\caption{The network architecture of our method. (a) SSO-PGA consists of $T$ iterative steps, where each iteration includes (b) the update of $\boldsymbol{\mathcal{H}}$ and (c) the update of $\boldsymbol{\mathcal{T}}$.
(d) The detailed network architecture of SSO-PGA, including the init module, MSFB block, and $Prox_{\phi}(\cdot)$ from left to right.}
\vspace{-10pt}
\label{fig:network}
\end{figure}
Based on the SSO-PGA, we update each variable alternately.

\textbf{Step 1:} $\boldsymbol{\mathcal{H}}$ can be updated as follows at the t-th iteration:
\begin{equation}
\begin{aligned}
\boldsymbol{\mathcal{H}}^t=Prox_{\phi}\left(
\boldsymbol{\mathcal{H}}^{t-1} \odot SSO_{\alpha_1}\left(\nabla \mathcal{E}(\boldsymbol{\mathcal{H}}^{t-1})\right)\right),
\end{aligned}
\label{eq:H1}
\end{equation}
where
\begin{equation}
\begin{aligned}
\mathcal{E}(\boldsymbol{\mathcal{H}}^{t-1}) = \|\boldsymbol{\mathcal{X}} -\boldsymbol{\mathcal{K}}\boldsymbol{\mathcal{H}}^{t-1}\|_F^2 +\gamma\|\boldsymbol{\mathcal{T}}^{t-1} - f(\boldsymbol{\mathcal{H}}^{t-1})\|_F^2,
\end{aligned}
\label{eq:H2}
\end{equation}
and
\begin{equation}
\begin{aligned}
\nabla \mathcal{E}(\boldsymbol{\mathcal{H}}^{t-1}) = 2\boldsymbol{\mathcal{K}}^{\top}(\boldsymbol{\mathcal{K}}\boldsymbol{\mathcal{H}}^{t-1} -\boldsymbol{\mathcal{X}})+2\gamma f^*\big(f(\boldsymbol{\mathcal{H}}^{t-1}) - \boldsymbol{\mathcal{T}}^{t-1}\big).
\end{aligned}
\label{eq:H3}
\end{equation}
Specifically, \( f^*(\cdot) \) is the subgradient of \( f(\cdot) \), and the proximal operator \( Prox_{\phi}(\cdot) \) is a deep network related to \( \phi(\cdot) \).
% Then, combined with the regularization term, the solution of Eq.~(\ref{eq:A1}) is:
% \begin{equation}
% \begin{aligned}
% \boldsymbol{A} = \boldsymbol{P}^{\top}\boldsymbol{P}\boldsymbol{A} =\boldsymbol{P}^{\top}\big[Soft\big((\boldsymbol{U}_1 - \boldsymbol{E}),\rho_1\gamma\big) +\boldsymbol{E}\big],
% \end{aligned}
% \label{eq:A4}
% \end{equation}
% where $\boldsymbol{U}_1 = \boldsymbol{A}^{t-1} - \rho_1\nabla \mathcal{E}(\boldsymbol{A}^{t-1})$, and $Soft(\cdot, \cdot)$ is the soft-thresholding operator.

\textbf{Step 2:} Similarly, we update $\boldsymbol{\mathcal{T}}$ as follows:
\begin{equation}
\begin{aligned}
\boldsymbol{\mathcal{T}}^t= 
\boldsymbol{\mathcal{T}}^{t-1} \odot SSO_{\alpha_2}\left(\nabla \mathcal{E}(\boldsymbol{\mathcal{T}}^{t-1})\right),
\end{aligned}
\label{eq:T1}
\end{equation}
where
\begin{equation}
\begin{aligned}
\mathcal{E}(\boldsymbol{\mathcal{T}}^{t-1}) = \beta\|\boldsymbol{\mathcal{Y}} -\boldsymbol{\mathcal{S}}\boldsymbol{\mathcal{T}}^{t-1}\|_F^2
+ \gamma\|\boldsymbol{\mathcal{T}}^{t-1} - f(\boldsymbol{\mathcal{H}}^{t})\|_F^2,
\end{aligned}
\label{eq:T2}
\end{equation}
and
\begin{equation}
\begin{aligned}
\nabla \mathcal{E}(\boldsymbol{\mathcal{T}}^{t-1}) = 2\beta\boldsymbol{\mathcal{S}}^{\top}(\boldsymbol{\mathcal{S}}\boldsymbol{\mathcal{T}}^{t-1} - \boldsymbol{\mathcal{Y}})+2\gamma (\boldsymbol{\mathcal{T}}^{t-1} - f(\boldsymbol{\mathcal{H}}^{t})).
\end{aligned}
\label{eq:T3}
\end{equation}

\subsection{Deep Unfolding Network}
This subsection unfolds the SSO-PGA framework into a deep network architecture. As shown in Fig.~\ref{fig:network}, the network begins with an initialization module, followed by multiple iterative stages. Each iteration comprises two submodules: one for updating $\boldsymbol{\mathcal{H}}$ and the other for updating $\boldsymbol{\mathcal{T}}$. In this formulation, the operators $\boldsymbol{\mathcal{K}}, \boldsymbol{\mathcal{K}}^\top, \boldsymbol{\mathcal{S}}, \boldsymbol{\mathcal{S}}^\top$ in the original optimization steps are replaced by a multi-scale spatial frequency feature extraction module (MSFB), while the functions $f(\cdot)$ and $f^*(\cdot)$ are implemented using an NAFBlock \cite{chen2022simple}. The proximal operator ${Prox}_\phi(\cdot)$ is modeled by a combination of multiple MSFBs and NAFBlocks \cite{chen2022simple}. Additionally, all hyperparameters in each iteration, including $\beta$, $\gamma$, $\alpha_1$, and $\alpha_2$, are learnable and passed through a Softplus function to enforce non-negativity. Finally, the final network output is obtained by adding the $\boldsymbol{\mathcal{H}}$ in the last iteration and the initial input, and an L1 loss is applied against the ground truth.

\section{Experiments}
\label{Experiments}
In this section, we conduct comprehensive experiments to validate the effectiveness of our method, including both numerical experiments and real-world vision experiments. 
% We focus on multispectral fusion here; please refer to the \textit{Appendix} for results on depth map reconstruction and thermal image super-resolution.
\subsection{Comparison with Traditional Proximal Gradient Algorithm}
\subsubsection{Numerical Experiments}
\begin{figure}[h!]
\centering
 \includegraphics[width = 0.99\textwidth]{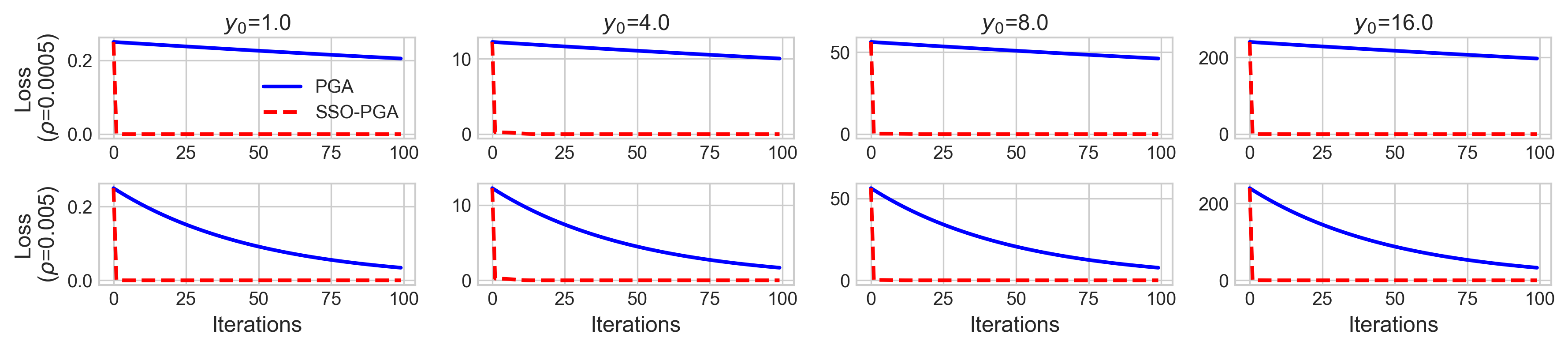}
\caption{Comparison of numerical simulation results for SSO-PGA and PGA on Problem I.}
\vspace{-15pt}
\label{fig:num1}
\end{figure}

\begin{figure}[h!]
\centering
 \includegraphics[width = 0.99\textwidth]{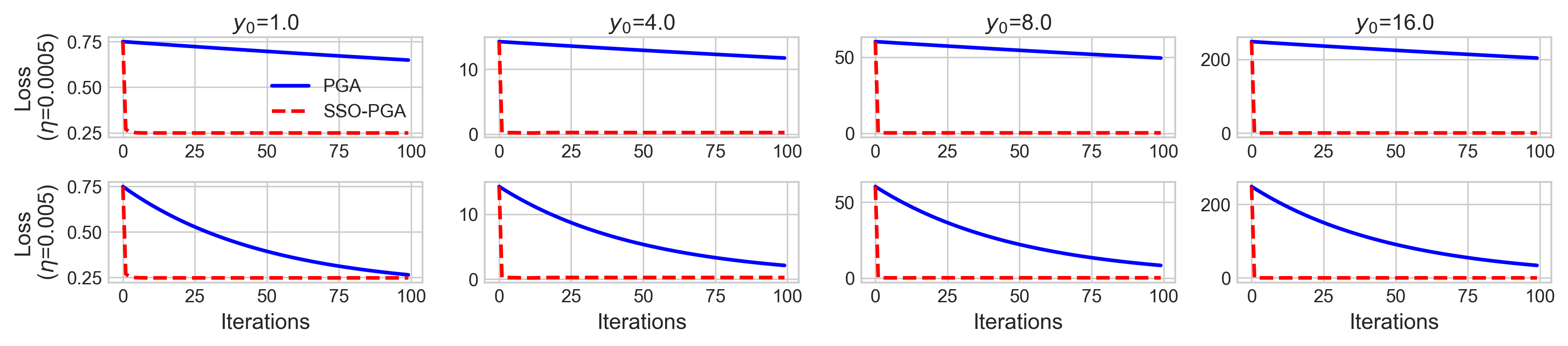}
\caption{Comparison of numerical simulation results for SSO-PGA and PGA on Problem II.}
\vspace{-15pt}
\label{fig:num2}
\end{figure}

In this subsection, we construct two convex optimization problems and perform numerical simulation experiments.
\begin{equation}
\begin{aligned}
&\min_{y} ({y}-0.5)^2,\quad &&(\text{Problem I}),\\
    &\min_{y} ({y}-0.5)^2+\frac{1}{2}|y|,\quad &&(\text{Problem II}).
\label{eq: numerical simulation experiments}
\end{aligned}
\end{equation}
We selected initial values of 1, 4, 8, and 16, with learning rates of 0.0005 and 0.005 (For additional experiments, please refer to the \textit{APPENDIX}). Fig.~\ref{fig:num1} and Fig.~\ref{fig:num2} show that our SSO-PGA has a clear advantage over PGA, which can be attributed to the benefits of our multiplicative update rule.

\subsubsection{Real-world Vision Experiments}
In this subsection, we construct a PGA baseline by replacing the SSO update rule in Eq.~(\ref{eq: ssoupdate}) with the traditional gradient descent formulation in Eq.~(\ref{eq:PGA}), while keeping all other components unchanged. We then conduct a comprehensive comparison with our proposed SSO-PGA.

\textbf{Performance Comparison.} To more intuitively verify the effectiveness of SSO, in addition to comparing SSO-PGA with PGA, we also replace the traditional gradient descent step in MDCUN \cite{yang2022memory} with our SSO-based update rule and compare it with the original version.
As shown in Tab.~\ref{tab:SSO_PGA_vs_PGA}, the SSO-enhanced models significantly outperform the traditional gradient descent models across all three datasets, demonstrating the superiority of the proposed SSO update mechanism.
\begin{table}[ht]
\centering
\caption{Quantitative comparison of traditional proximal gradient algorithm and SSO-enhanced proximal gradient algorithm on three datasets: WV3, QB, and GF2. The better results are in \textbf{bold}.}
\renewcommand{\arraystretch}{1.2}
\setlength{\tabcolsep}{3.6pt}
\resizebox{\textwidth}{!}{%
\begin{tabular}{l|cccc|cccc|cccc}
\toprule
\multirow{2}{*}{Methods} & \multicolumn{4}{c|}{\textbf{WV3}} & \multicolumn{4}{c|}{\textbf{QB}} & \multicolumn{4}{c}{\textbf{GF2}} \\
 & PSNR↑ & SAM↓ & ERGAS↓ & Q8↑ & PSNR↑ & SAM↓ & ERGAS↓ & Q4↑ & PSNR↑ & SAM↓ & ERGAS↓ & Q4↑ \\
\midrule
MDCUN \cite{yang2022memory}    & 37.973  & 3.298 & 2.479  & \textbf{0.909} & 36.178 & \textbf{4.963} & 4.698 & 0.915  & 41.138 & 0.870 & 0.815 & 0.974  \\
SSO-MDCUN   & \textbf{38.135} & \textbf{3.222} & \textbf{2.437} & \textbf{0.909} & \textbf{36.462} & 5.007 & \textbf{4.527} & \textbf{0.917} & \textbf{41.626} & \textbf{0.869} & \textbf{0.783} & \textbf{0.976}  \\
\midrule
PGA     &  39.145 & 2.925 & 2.129 & 0.918 & 38.628 & 4.430 & 3.557 & 0.937 & 43.411 & 0.697 & 0.615 & 0.982 \\
SSO-PGA  & \textbf{39.358} & \textbf{2.823} & \textbf{2.078} & \textbf{0.921} & \textbf{38.807} & \textbf{4.312} & \textbf{3.493} & \textbf{0.938} & \textbf{44.005} & \textbf{0.660} & \textbf{0.574} & \textbf{0.985} \\
\bottomrule
\end{tabular}
}
\label{tab:SSO_PGA_vs_PGA}
\end{table}

\begin{minipage}[t]{0.50\textwidth}
\vspace{-10pt}
\textbf{Convergence Behavior.}
Fig.~\ref{fig:psnr-iter} illustrates the convergence behavior of SSO-PGA and PGA under varying numbers of iterations. As observed, both methods perform comparably at the first iteration, because the model at this stage mainly behaves like a deep network, and the iterative formulation has not yet taken effect. However, with just two iterations, SSO-PGA already surpasses the three-iteration performance of PGA. By the third iteration, SSO-PGA exceeds the best performance achieved by PGA.
Notably, at higher iteration counts, PGA exhibits signs of performance degradation, whereas SSO-PGA continues to improve steadily. This indicates that SSO-PGA not only converges faster but is also more robust against falling into poor local minima.
\end{minipage}
\hfill
\begin{minipage}[t]{0.46\textwidth}
\vspace{0pt}
\centering
\includegraphics[width=\linewidth]{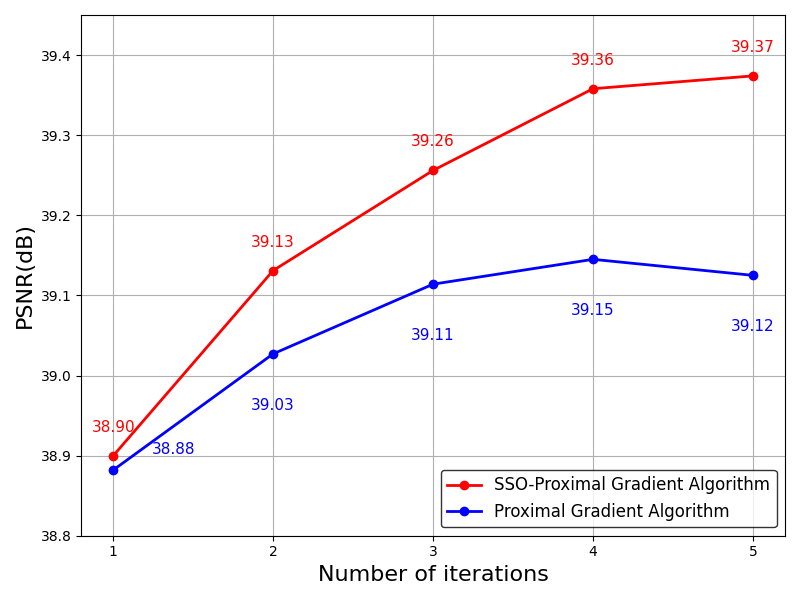}
\captionof{figure}{PSNR comparison of SSO-PGA and PGA over iterations on the WV3 dataset.}
\label{fig:psnr-iter}
\end{minipage}
\begin{table}[ht]
\centering
\caption{Quantitative comparison of SSO-PGA and PGA  on the WV3 reduced-resolution dataset with varying parameter initialization settings. The better results are in \textbf{bold}.}
\renewcommand{\arraystretch}{1.2}
\setlength{\tabcolsep}{3.6pt}
\resizebox{\textwidth}{!}{%
\begin{tabular}{l|cccc|cccc|cccc}
\toprule
\multirow{2}{*}{Parameter} & \multicolumn{4}{c|}{$\alpha ,\rho$ = 0.01} & \multicolumn{4}{c|}{$\alpha ,\rho$ = 0.1} & \multicolumn{4}{c}{$\alpha ,\rho$ = 0.5} \\
 & PSNR↑ & SAM↓ & ERGAS↓ & Q8↑ & PSNR↑ & SAM↓ & ERGAS↓ & Q8↑ & PSNR↑ & SAM↓ & ERGAS↓ & Q8↑ \\
\midrule
PGA    & 39.116 & 2.948 & 2.143 & 0.919 & 39.145 & 2.925 & 2.129 & 0.918 & 39.063 & 2.923 & 2.149 & 0.918 \\
SSO-PGA     & \textbf{39.223} & \textbf{2.859} & \textbf{2.119} & \textbf{0.920} & \textbf{39.191} & \textbf{2.863} & \textbf{2.116} & \textbf{0.920} & \textbf{39.283} & \textbf{2.847} & \textbf{2.095} & \textbf{0.920} \\
\midrule
\multirow{2}{*}{Parameter} & \multicolumn{4}{c|}{$\alpha ,\rho$ = 1.0} & \multicolumn{4}{c|}{$\alpha ,\rho$ = 3.0} & \multicolumn{4}{c}{$\alpha ,\rho$ = 5.0} \\
 & PSNR↑ & SAM↓ & ERGAS↓ & Q8↑ & PSNR↑ & SAM↓ & ERGAS↓ & Q8↑ & PSNR↑ & SAM↓ & ERGAS↓ & Q8↑ \\
\midrule
PGA    & 38.907	&3.017	&2.195	&0.917 & 23.162	&31.146	&14.431	&0.490	&7.678	&46.171	&110.511	&0.011 \\
SSO-PGA     & \textbf{39.358} & \textbf{2.823} & \textbf{2.078} & \textbf{0.921} & \textbf{39.225} & \textbf{2.857} & \textbf{2.108} & \textbf{0.921} & \textbf{39.171} & \textbf{2.876} & \textbf{2.123} & \textbf{0.919} \\
\bottomrule
\end{tabular}
}
\label{tab:para}
\end{table}

\textbf{Parameter Sensitivity.}
Both SSO-PGA and PGA involve two hyperparameters during the update process: the sliding factor $\alpha_1,\alpha_2$ for SSO-PGA and the step size $\rho_1,\rho_2$ for PGA. As noted in our deep unfolding network, these hyperparameters are learnable. Here, we assign multiple initial values to $\alpha$ and $\rho$ to evaluate the sensitivity of SSO-PGA and PGA to the hyperparameter.
Tab.~\ref{tab:para} shows that PGA achieves its best performance when $\rho = 0.1$, and suboptimal results when $\rho = 0.01$. In contrast, SSO-PGA consistently performs well across all initial values. Notably, when the hyperparameter is set to relatively large values (e.g., $3.0$ or $5.0$), PGA fails to converge, whereas SSO-PGA still delivers strong performance. These results further confirm the robustness and stability of the proposed SSO-PGA framework.

\subsection{Comparison with SOTAs}
\subsubsection{Multispectral Image Fusion}
\textbf{Datasets and Setting.} We conducted experiments on three datasets consisting of satellite images captured by WorldView-3 (WV3), QuickBird (QB), and GaoFen-2 (GF2), provided by the PanCollection repository \cite{deng2022machine}. We evaluate our method using a set of widely used performance metrics. For reduced-resolution data, we use PSNR, SAM \cite{boardman1993automating}, ERGAS \cite{wald2002data}, and Q4/Q8 \cite{garzelli2009hypercomplex}.

\begin{table}[ht]
\centering
\caption{Quantitative comparison for multispectral image fusion on reduced-resolution datasets: WV3, QB, and GF2. The best results are in \textbf{bold} and the second-best values are \underline{underlined}.}
\renewcommand{\arraystretch}{1.2}
\setlength{\tabcolsep}{3.6pt}
\resizebox{\textwidth}{!}{%
\begin{tabular}{l|cccc|cccc|cccc}
\toprule
\multirow{2}{*}{Methods} & \multicolumn{4}{c|}{\textbf{WV3}} & \multicolumn{4}{c|}{\textbf{QB}} & \multicolumn{4}{c}{\textbf{GF2}} \\
% \cmidrule(lr){2-5} \cmidrule(lr){6-9} \cmidrule(lr){10-13} 
 & PSNR↑ & SAM↓ & ERGAS↓ & Q8↑ & PSNR↑ & SAM↓ & ERGAS↓ & Q4↑ & PSNR↑ & SAM↓ & ERGAS↓ & Q4↑ \\
\midrule
MTF-GLP-FS \cite{vivone2018full} & 32.963 & 5.316 & 4.700 & 0.833 & 32.709 & 7.792 & 7.373 & 0.835 & 35.540 & 1.655 & 1.589 & 0.897 \\
BDSD-PC \cite{vivone2019robust}   & 32.970 & 5.428 & 4.697 & 0.829 & 32.550 & 8.085 & 7.513 & 0.831 & 35.180 & 1.681 & 1.667 & 0.892 \\
TV \cite{palsson2013new}       & 32.381 & 5.692 & 4.855 & 0.795 & 32.136 & 7.510 & 7.690 & 0.821 & 35.237 & 1.911 & 1.737 & 0.907 \\
PNN \cite{masi2016pansharpening}       & 37.313 & 3.677 & 2.681 & 0.893 & 36.942 & 5.181 & 4.468 & 0.918 & 39.071 & 1.048 & 1.057 & 0.960 \\
PanNet \cite{yang2017pannet}    & 37.346 & 3.613 & 2.664 & 0.891 & 34.678 & 5.767 & 5.859 & 0.885 & 40.243 & 0.997 & 0.919 & 0.967 \\
DiCNN \cite{he2019pansharpening}     & 37.390 & 3.592 & 2.672 & 0.900 & 35.781 & 5.367 & 5.133 & 0.904 & 38.906 & 1.053 & 1.081 & 0.959 \\
FusionNet \cite{deng2020detail} & 38.047 & 3.324 & 2.465 & 0.904 & 37.540 & 4.904 & 4.156 & 0.925 & 39.639 & 0.974 & 0.988 & 0.964 \\
MDCUN \cite{yang2022memory}    & 37.973  & 3.298 & 2.479  & 0.909 & 36.178 & 4.963 & 4.698 & 0.915  & 41.138 & 0.870 & 0.815 & 0.974  \\
LAGNet \cite{jin2022lagconv}    & 38.592 & 3.103 & 2.291 & 0.910 & 38.209 & 4.534 & 3.812 & 0.934 & 42.735 & 0.786 & 0.687 & 0.980 \\
LGPNet \cite{zhao2023lgpconv}    & 38.147 & 3.270 & 2.422 & 0.902 & 36.443 & 4.954 & 4.777 & 0.915 & 41.843 & 0.845 & 0.765 & 0.976 \\
U2Net \cite{peng2023u2net} & 39.117 & \underline{2.888} & 2.149 & \underline{0.920} & 38.065 & 4.642 & 3.987 & 0.931 & 43.379 & 0.714 & 0.632 & 0.981\\
CANNet \cite{duan2024content}    &  39.003 & 2.941 &  2.174 & \underline{0.920} & \underline{38.488} & 4.496 & \underline{3.698} & \underline{0.937} & 43.496 & 0.707 & 0.630 & \underline{0.983} \\
PanMamba \cite{he2025pan}  & 39.012 & 2.913 & 2.184 & \underline{0.920} & 37.356 & 4.625 & 4.277 & 0.929 & 42.907 & 0.743 & 0.684 & 0.982 \\
ADWM \cite{huang2025general}    &  \underline{39.170} & 2.913 &  \underline{2.145} & \textbf{0.921} & 38.466 & \underline{4.450} & 3.705 & \underline{0.937} & \underline{43.884} & \underline{0.672} & \underline{0.597} & \textbf{0.985} \\
SSO-PGA (ours)     & \textbf{39.358} & \textbf{2.823} & \textbf{2.078} & \textbf{0.921} & \textbf{38.807} & \textbf{4.312} & \textbf{3.493} & \textbf{0.938} & \textbf{44.005} & \textbf{0.660} & \textbf{0.574} & \textbf{0.985} \\
\bottomrule
\end{tabular}
}
\label{tab:quantitative_comparison}
\end{table}

\begin{figure}
\centering
 \includegraphics[width = 0.99\textwidth]{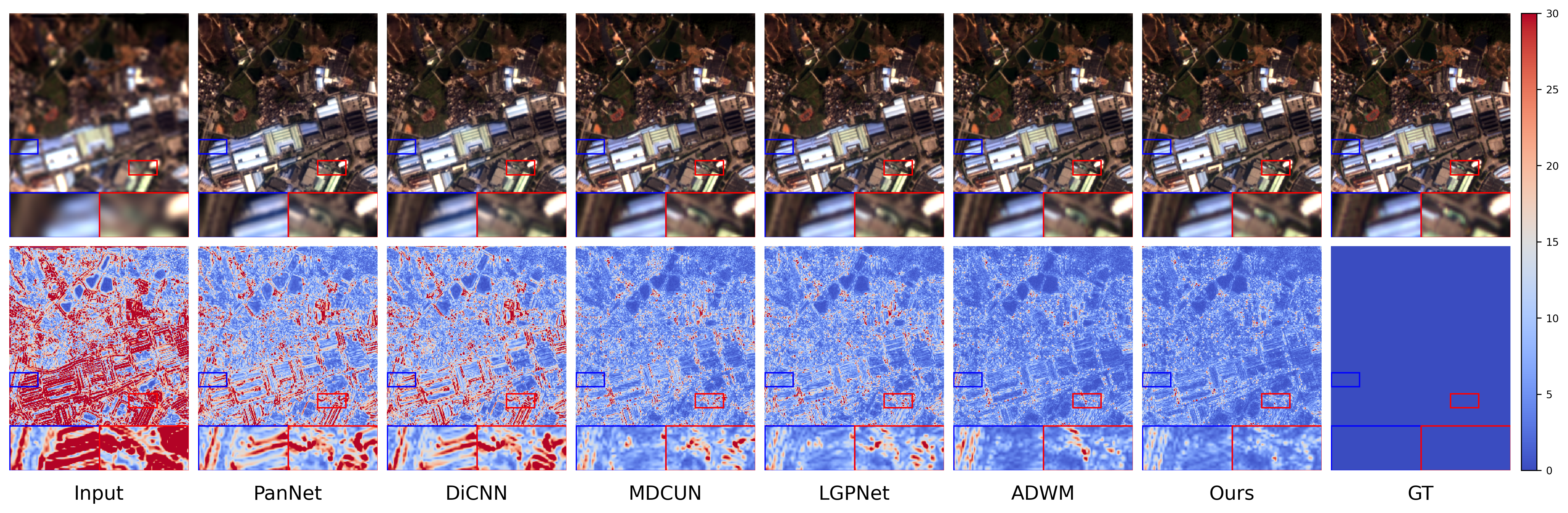}
\caption{Visual comparison (the first row) and the corresponding error map (the second row) of our method and some representative methods on the GF2 reduced-resolution dataset.}
\label{fig:visual1}
\end{figure}

\textbf{Experimental Results.} As shown in Tab.~\ref{tab:quantitative_comparison}, our proposed SSO-PGA consistently achieves the best results across all datasets compared to other methods. Specifically, in terms of PSNR, our method outperforms the second-best method by 0.188 dB, 0.319 dB, and 0.121 dB on the WV3, QB, and GF2 datasets, respectively. These consistent improvements validate the effectiveness of our deep unfolding framework. Furthermore, Fig.~\ref{fig:visual1} presents a qualitative visual comparison of the GF2 dataset against several representative methods. Our method produces reconstructions closer to the ground truth with lower residuals, further highlighting its superiority.

\subsubsection{Flash Guided Non-Flash Image Denoising}
\textbf{Datasets and Setting.} 
Following the experimental protocol in recent studies \cite{10323520, xu2024laplacian}, we used the following datasets for training and testing: the Flash and Ambient Illuminations Dataset (FAID) \cite{aksoy2018dataset} and the Multi-Illumination Dataset (MID) \cite{murmann2019dataset}. We added varying levels of Gaussian noise to the non-flash images in each dataset and used PSNR as the evaluation metric. 

\textbf{Experimental Results.} 
As shown in Tab.~\ref{tab:flash_results}, our method outperforms the others on MID and FAID datasets. This not only highlights the performance of our method but also demonstrates its versatility and generalization capabilities across various tasks. It's worth noting that although our method's performance is on par with DeepM$^2$CDL \cite{10323520}, our method has a parameter count of just 2.90M, which is significantly smaller than DeepM$^2$CDL \cite{10323520}'s 421.14M. This highlights the lightweight and efficient nature of our approach, as it minimizes computational overhead while maintaining comparable performance.

\begin{table}[htbp]
 \centering
 \caption{Quantitative comparison for flash guided non-flash image denoising in terms of PSNR (dB) on MID and FAID datasets. The best results are in \textbf{bold} and the second-best values are \underline{underlined}.}
 \resizebox{\textwidth}{!}{%
 \begin{tabular}{ll|ccc|ccc}
 \toprule
 \multicolumn{2}{l|}{\multirow{2}{*}{Methods}} & \multicolumn{3}{c|}{MID} & \multicolumn{3}{c}{FAID} \\
 % \cmidrule(lr){2-4} \cmidrule(lr){5-7} \cmidrule(lr){8-10}
 & & $\sigma=25$ & $\sigma=50$ & $\sigma=75$ & $\sigma=25$ & $\sigma=50$ & $\sigma=75$ \\
 \midrule
 DnCNN \cite{zhang2017beyond} & & 34.57 & 32.69 & 31.26 & 35.38 & 31.94 & 30.08 \\
 DJFR \cite{li2019joint} & & 37.03 & 32.96 & 31.84 & 33.76 & 30.61 & 28.92 \\
 CUNet \cite{deng2020deep} & & 34.61 & 32.39 & 31.18 & 35.86 & 33.05 & 31.30 \\
 UMGF \cite{shi2021unsharp} & & 38.18& 35.84 & 34.30 & 34.52 & 31.81 & 30.43 \\
 MN \cite{xu2022model} & & 39.51 & 37.01 & 35.50 & 36.15 & 33.34 & 31.83 \\
 FGDNet \cite{sheng2022frequency} & & 38.38 & 35.88 & 34.39 & 34.99 & 32.15 & 30.81 \\
 RIDFhF \cite{oh2023robust} & & 38.31 & 35.33 & 33.74 & 36.25 & 33.48 & 31.92 \\
 DeepM$^2$CDL \cite{10323520} & (Para: 421.14M)& \underline{39.67} & \underline{37.61} & \textbf{36.28} & \underline{36.86} & \textbf{34.43} & \textbf{32.95} \\
 SSO-PGA (ours) & (Para: 2.90M)& \textbf{39.84} & \textbf{37.66} & \underline{35.71} & \textbf{36.88} & \underline{34.12} & \underline{32.92} \\
    \bottomrule
  \end{tabular}
  }
  \label{tab:flash_results}
\end{table}

\subsection{Ablation Study}
\begin{minipage}[t]{0.48\textwidth}
We conduct a comprehensive ablation study on the SSO-PGA network. First, we remove the ${Prox}_\phi(\cdot)$ module to construct the variant V-1. Then, we individually remove the standard convolution, dilated convolution, and frequency-domain convolution from the MSFB to construct variants V-2, V-3, and V-4, respectively. The results in Tab.~\ref{tab:abaltion} show that SSO-PGA outperforms variant V-1, which demonstrates the importance
\end{minipage}
\begin{minipage}[t]{0.5\textwidth}
\centering
\captionsetup{type=table}
\captionof{table}{Ablation Study of different variants.}
\setlength{\tabcolsep}{6pt}
\scriptsize
\begin{tabular}{l|cccc}
\toprule
 Variant & PSNR↑ & SAM↓ & ERGAS↓ & Q8↑ \\
\midrule
V-1 & 38.194 & 3.176 & 2.396 & 0.911 \\
V-2 & 39.301 & 2.849 & 2.088 & 0.920 \\
V-3 & 39.190 & 2.868 & 2.108 & 0.919 \\
V-4 & 39.236 & 2.854 & 2.106 & \textbf{0.921} \\
ours    & \textbf{39.358} & \textbf{2.823} & \textbf{2.078} & \textbf{0.921} \\
\bottomrule
\end{tabular}
\label{tab:abaltion}
\end{minipage}
 of the deep prior. Furthermore, the superiority of SSO-PGA over V-2, V-3, and V-4 verifies that each branch in the MSFB module is indispensable and plays a critical role in enabling comprehensive information fusion.

\section{Conclusion}
This paper proposes SSO-PGA, a novel multiplicative proximal gradient algorithm enhanced by Sliding Sigmoid Operator, which improves stability and adaptivity. We replace the traditional gradient descent step with a learnable sigmoid-based operator, which inherently enforces non-negativity and boundedness. SSO-PGA is formulated for multi-modal restoration. We then iteratively solve the model and further unfold it into a deep neural network. Both numerical and real-world experiments verify the superiority of SSO-PGA and its significant improvements in accuracy and convergence speed over conventional PGA. Future work will focus on analyzing the theoretical convergence rate of SSO-PGA and extending its application to broader vision tasks.


\bibliographystyle{unsrt}
\bibliography{main}

\begin{thebibliography}{10}

\bibitem{ruder2016overview}
Sebastian Ruder.
\newblock An overview of gradient descent optimization algorithms.
\newblock {\em arXiv preprint arXiv:1609.04747}, 2016.

\bibitem{li2021decentralized}
Huaqing Li, Jinhui Hu, Liang Ran, Zheng Wang, Qingguo L{\"u}, Zhenyuan Du, and Tingwen Huang.
\newblock Decentralized dual proximal gradient algorithms for non-smooth constrained composite optimization problems.
\newblock {\em IEEE Transactions on Parallel and Distributed Systems}, 32(10):2594--2605, 2021.

\bibitem{goldfarb2012fast}
Donald Goldfarb and Shiqian Ma.
\newblock Fast multiple-splitting algorithms for convex optimization.
\newblock {\em SIAM Journal on Optimization}, 22(2):533--556, 2012.

\bibitem{li2015accelerated}
Huan Li and Zhouchen Lin.
\newblock Accelerated proximal gradient methods for nonconvex programming.
\newblock {\em Advances in neural information processing systems}, 28, 2015.

\bibitem{salim2020wasserstein}
Adil Salim, Anna Korba, and Giulia Luise.
\newblock The wasserstein proximal gradient algorithm.
\newblock {\em Advances in Neural Information Processing Systems}, 33:12356--12366, 2020.

\bibitem{boyd2011distributed}
Stephen Boyd, Neal Parikh, Eric Chu, Borja Peleato, Jonathan Eckstein, et~al.
\newblock Distributed optimization and statistical learning via the alternating direction method of multipliers.
\newblock {\em Foundations and Trends{\textregistered} in Machine learning}, 3(1):1--122, 2011.

\bibitem{hong2017linear}
Mingyi Hong and Zhi-Quan Luo.
\newblock On the linear convergence of the alternating direction method of multipliers.
\newblock {\em Mathematical Programming}, 162(1):165--199, 2017.

\bibitem{eckstein1992douglas}
Jonathan Eckstein and Dimitri~P Bertsekas.
\newblock On the douglas—rachford splitting method and the proximal point algorithm for maximal monotone operators.
\newblock {\em Mathematical programming}, 55(1):293--318, 1992.

\bibitem{patrinos2014douglas}
Panagiotis Patrinos, Lorenzo Stella, and Alberto Bemporad.
\newblock Douglas-rachford splitting: Complexity estimates and accelerated variants.
\newblock In {\em 53rd IEEE Conference on Decision and Control}, pages 4234--4239. IEEE, 2014.

\bibitem{chambolle2011first}
Antonin Chambolle and Thomas Pock.
\newblock A first-order primal-dual algorithm for convex problems with applications to imaging.
\newblock {\em Journal of mathematical imaging and vision}, 40(1):120--145, 2011.

\bibitem{dai2024proximal}
Yutong Dai, Xiaoyi Qu, and Daniel~P Robinson.
\newblock A proximal-gradient method for constrained optimization.
\newblock {\em arXiv preprint arXiv:2404.07460}, 2024.

\bibitem{laude2025anisotropic}
Emanuel Laude and Panagiotis Patrinos.
\newblock Anisotropic proximal gradient.
\newblock {\em Mathematical Programming}, pages 1--45, 2025.

\bibitem{li2019decentralized}
Zhi Li, Wei Shi, and Ming Yan.
\newblock A decentralized proximal-gradient method with network independent step-sizes and separated convergence rates.
\newblock {\em IEEE Transactions on Signal Processing}, 67(17):4494--4506, 2019.

\bibitem{iutzeler2018proximal}
Franck Iutzeler and J{\'e}r{\^o}me Malick.
\newblock On the proximal gradient algorithm with alternated inertia.
\newblock {\em Journal of Optimization Theory and Applications}, 176(3):688--710, 2018.

\bibitem{si2024riemannian}
Wutao Si, P-A Absil, Wen Huang, Rujun Jiang, and Simon Vary.
\newblock A riemannian proximal newton method.
\newblock {\em SIAM Journal on Optimization}, 34(1):654--681, 2024.

\bibitem{keys2019proximal}
Kevin~L Keys, Hua Zhou, and Kenneth Lange.
\newblock Proximal distance algorithms: Theory and practice.
\newblock {\em Journal of Machine Learning Research}, 20(66):1--38, 2019.

\bibitem{malitsky2024adaptive}
Yura Malitsky and Konstantin Mishchenko.
\newblock Adaptive proximal gradient method for convex optimization.
\newblock {\em Advances in Neural Information Processing Systems}, 37:100670--100697, 2024.

\bibitem{antonello2018proximal}
Niccol{\`o} Antonello, Lorenzo Stella, Panagiotis Patrinos, and Toon Van~Waterschoot.
\newblock Proximal gradient algorithms: Applications in signal processing.
\newblock {\em arXiv preprint arXiv:1803.01621}, 2018.

\bibitem{yao2025low}
Xi~Yao and Wei Dai.
\newblock A low-rank projected proximal gradient method for spectral compressed sensing.
\newblock {\em IEEE Transactions on Signal Processing}, 2025.

\bibitem{shen2011accelerated}
Zuowei Shen, Kim-Chuan Toh, and Sangwoon Yun.
\newblock An accelerated proximal gradient algorithm for frame-based image restoration via the balanced approach.
\newblock {\em SIAM Journal on Imaging Sciences}, 4(2):573--596, 2011.

\bibitem{wei2022deep}
Xinyi Wei, Hans Van~Gorp, Lizeth Gonzalez-Carabarin, Daniel Freedman, Yonina~C Eldar, and Ruud~JG van Sloun.
\newblock Deep unfolding with normalizing flow priors for inverse problems.
\newblock {\em IEEE Transactions on Signal Processing}, 70:2962--2971, 2022.

\bibitem{mou2022deep}
Chong Mou, Qian Wang, and Jian Zhang.
\newblock Deep generalized unfolding networks for image restoration.
\newblock In {\em Proceedings of the IEEE/CVF Conference on Computer Vision and Pattern Recognition}, pages 17399--17410, 2022.

\bibitem{mardani2018neural}
Morteza Mardani, Qingyun Sun, David Donoho, Vardan Papyan, Hatef Monajemi, Shreyas Vasanawala, and John Pauly.
\newblock Neural proximal gradient descent for compressive imaging.
\newblock {\em Advances in Neural Information Processing Systems}, 31, 2018.

\bibitem{xin2024rethinking}
Bingyu Xin, Meng Ye, Leon Axel, and Dimitris~N Metaxas.
\newblock Rethinking deep unrolled model for accelerated mri reconstruction.
\newblock In {\em European Conference on Computer Vision}, pages 164--181. Springer, 2024.

\bibitem{deng2018variational}
Liang-Jian Deng, Gemine Vivone, Weihong Guo, Mauro Dalla~Mura, and Jocelyn Chanussot.
\newblock A variational pansharpening approach based on reproducible kernel hilbert space and heaviside function.
\newblock {\em IEEE Transactions on Image Processing}, 27(9):4330--4344, 2018.

\bibitem{farahmand2024robust}
Salar Farahmand-Tabar, Fahimeh Abdollahi, and Masoud Fatemi.
\newblock Robust conjugate gradient methods for non-smooth convex optimization and image processing problems.
\newblock In {\em Handbook of formal optimization}, pages 19--43. Springer, 2024.

\bibitem{nan2020deep}
Yuesong Nan and Hui Ji.
\newblock Deep learning for handling kernel/model uncertainty in image deconvolution.
\newblock In {\em Proceedings of the IEEE/CVF Conference on Computer Vision and Pattern Recognition}, pages 2388--2397, 2020.

\bibitem{he2014new}
Xiyan He, Laurent Condat, Jos{\'e}~M Bioucas-Dias, Jocelyn Chanussot, and Junshi Xia.
\newblock A new pansharpening method based on spatial and spectral sparsity priors.
\newblock {\em IEEE Transactions on Image Processing}, 23(9):4160--4174, 2014.

\bibitem{palsson2013new}
Frosti Palsson, Johannes~R Sveinsson, and Magnus~O Ulfarsson.
\newblock A new pansharpening algorithm based on total variation.
\newblock {\em IEEE Geoscience and Remote Sensing Letters}, 11(1):318--322, 2013.

\bibitem{beck2009fast}
Amir Beck and Marc Teboulle.
\newblock A fast iterative shrinkage-thresholding algorithm for linear inverse problems.
\newblock {\em SIAM journal on imaging sciences}, 2(1):183--202, 2009.

\bibitem{lee2000algorithms}
Daniel Lee and H~Sebastian Seung.
\newblock Algorithms for non-negative matrix factorization.
\newblock {\em Advances in Neural Information Processing Systems}, 13, 2000.

\bibitem{nesterov2013introductory}
Yurii Nesterov.
\newblock {\em Introductory lectures on convex optimization: A basic course}, volume~87.
\newblock Springer Science \& Business Media, 2013.

\bibitem{chen2022simple}
Liangyu Chen, Xiaojie Chu, Xiangyu Zhang, and Jian Sun.
\newblock Simple baselines for image restoration.
\newblock In {\em European Conference on Computer Vision}, pages 17--33. Springer, 2022.

\bibitem{yang2022memory}
Gang Yang, Man Zhou, Keyu Yan, Aiping Liu, Xueyang Fu, and Fan Wang.
\newblock Memory-augmented deep conditional unfolding network for pan-sharpening.
\newblock In {\em Proceedings of the IEEE/CVF Conference on Computer Vision and Pattern Recognition}, pages 1788--1797, 2022.

\bibitem{deng2022machine}
Liang-Jian Deng, Gemine Vivone, Mercedes~E Paoletti, Giuseppe Scarpa, Jiang He, Yongjun Zhang, Jocelyn Chanussot, and Antonio Plaza.
\newblock Machine learning in pansharpening: A benchmark, from shallow to deep networks.
\newblock {\em IEEE Geoscience and Remote Sensing Magazine}, 10(3):279--315, 2022.

\bibitem{boardman1993automating}
Joseph~W Boardman.
\newblock Automating spectral unmixing of aviris data using convex geometry concepts.
\newblock In {\em JPL, Summaries of the 4th Annual JPL Airborne Geoscience Workshop. Volume 1: AVIRIS Workshop}, 1993.

\bibitem{wald2002data}
Lucien Wald.
\newblock {\em Data fusion: definitions and architectures: fusion of images of different spatial resolutions}.
\newblock Presses des MINES, 2002.

\bibitem{garzelli2009hypercomplex}
Andrea Garzelli and Filippo Nencini.
\newblock Hypercomplex quality assessment of multi/hyperspectral images.
\newblock {\em IEEE Geoscience and Remote Sensing Letters}, 6(4):662--665, 2009.

\bibitem{vivone2018full}
Gemine Vivone, Rocco Restaino, and Jocelyn Chanussot.
\newblock Full scale regression-based injection coefficients for panchromatic sharpening.
\newblock {\em IEEE Transactions on Image Processing}, 27(7):3418--3431, 2018.

\bibitem{vivone2019robust}
Gemine Vivone.
\newblock Robust band-dependent spatial-detail approaches for panchromatic sharpening.
\newblock {\em IEEE transactions on Geoscience and Remote Sensing}, 57(9):6421--6433, 2019.

\bibitem{masi2016pansharpening}
Giuseppe Masi, Davide Cozzolino, Luisa Verdoliva, and Giuseppe Scarpa.
\newblock Pansharpening by convolutional neural networks.
\newblock {\em Remote Sensing}, 8(7):594, 2016.

\bibitem{yang2017pannet}
Junfeng Yang, Xueyang Fu, Yuwen Hu, Yue Huang, Xinghao Ding, and John Paisley.
\newblock Pannet: A deep network architecture for pan-sharpening.
\newblock In {\em Proceedings of the IEEE International Conference on Computer Vision}, pages 5449--5457, 2017.

\bibitem{he2019pansharpening}
Lin He, Yizhou Rao, Jun Li, Jocelyn Chanussot, Antonio Plaza, Jiawei Zhu, and Bo~Li.
\newblock Pansharpening via detail injection based convolutional neural networks.
\newblock {\em IEEE Journal of Selected Topics in Applied Earth Observations and Remote Sensing}, 12(4):1188--1204, 2019.

\bibitem{deng2020detail}
Liang-Jian Deng, Gemine Vivone, Cheng Jin, and Jocelyn Chanussot.
\newblock Detail injection-based deep convolutional neural networks for pansharpening.
\newblock {\em IEEE Transactions on Geoscience and Remote Sensing}, 59(8):6995--7010, 2020.

\bibitem{jin2022lagconv}
Zi-Rong Jin, Tian-Jing Zhang, Tai-Xiang Jiang, Gemine Vivone, and Liang-Jian Deng.
\newblock Lagconv: Local-context adaptive convolution kernels with global harmonic bias for pansharpening.
\newblock In {\em Proceedings of the AAAI Conference on Artificial Intelligence}, volume~36, pages 1113--1121, 2022.

\bibitem{zhao2023lgpconv}
Chen-Yu Zhao, Tian-Jing Zhang, Ran Ran, Zhi-Xuan Chen, and Liang-Jian Deng.
\newblock Lgpconv: Learnable gaussian perturbation convolution for lightweight pansharpening.
\newblock In {\em IJCAI}, pages 4647--4655, 2023.

\bibitem{peng2023u2net}
Siran Peng, Chenhao Guo, Xiao Wu, and Liang-Jian Deng.
\newblock U2net: A general framework with spatial-spectral-integrated double u-net for image fusion.
\newblock In {\em Proceedings of the 31st ACM International Conference on Multimedia}, pages 3219--3227, 2023.

\bibitem{duan2024content}
Yule Duan, Xiao Wu, Haoyu Deng, and Liang-Jian Deng.
\newblock Content-adaptive non-local convolution for remote sensing pansharpening.
\newblock In {\em Proceedings of the IEEE/CVF Conference on Computer Vision and Pattern Recognition}, pages 27738--27747, 2024.

\bibitem{he2025pan}
Xuanhua He, Ke~Cao, Jie Zhang, Keyu Yan, Yingying Wang, Rui Li, Chengjun Xie, Danfeng Hong, and Man Zhou.
\newblock Pan-mamba: Effective pan-sharpening with state space model.
\newblock {\em Information Fusion}, 115:102779, 2025.

\bibitem{huang2025general}
Jie Huang, Haorui Chen, Jiaxuan Ren, Siran Peng, and Liangjian Deng.
\newblock A general adaptive dual-level weighting mechanism for remote sensing pansharpening.
\newblock {\em arXiv preprint arXiv:2503.13214}, 2025.

\bibitem{10323520}
Xin Deng, Jingyi Xu, Fangyuan Gao, Xiancheng Sun, and Mai Xu.
\newblock Deep$\mathrm {M^{2}}$cdl: Deep multi-scale multi-modal convolutional dictionary learning network.
\newblock {\em IEEE Transactions on Pattern Analysis and Machine Intelligence}, 46(5):2770--2787, 2024.

\bibitem{xu2024laplacian}
Jingyi Xu, Xin Deng, Chenxiao Zhang, Shengxi Li, and Mai Xu.
\newblock Laplacian gradient consistency prior for flash guided non-flash image denoising.
\newblock {\em IEEE Transactions on Image Processing}, 2024.

\bibitem{aksoy2018dataset}
Yagiz Aksoy, Changil Kim, Petr Kellnhofer, Sylvain Paris, Mohamed Elgharib, Marc Pollefeys, and Wojciech Matusik.
\newblock A dataset of flash and ambient illumination pairs from the crowd.
\newblock In {\em Proceedings of the European Conference on Computer Vision (ECCV)}, pages 634--649, 2018.

\bibitem{murmann2019dataset}
Lukas Murmann, Michael Gharbi, Miika Aittala, and Fredo Durand.
\newblock A dataset of multi-illumination images in the wild.
\newblock In {\em Proceedings of the IEEE/CVF International Conference on Computer Vision}, pages 4080--4089, 2019.

\bibitem{zhang2017beyond}
Kai Zhang, Wangmeng Zuo, Yunjin Chen, Deyu Meng, and Lei Zhang.
\newblock Beyond a gaussian denoiser: Residual learning of deep cnn for image denoising.
\newblock {\em IEEE transactions on image processing}, 26(7):3142--3155, 2017.

\bibitem{li2019joint}
Yijun Li, Jia-Bin Huang, Narendra Ahuja, and Ming-Hsuan Yang.
\newblock Joint image filtering with deep convolutional networks.
\newblock {\em IEEE Transactions on Pattern Analysis and Machine Intelligence}, 41(8):1909--1923, 2019.

\bibitem{deng2020deep}
Xin Deng and Pier~Luigi Dragotti.
\newblock Deep convolutional neural network for multi-modal image restoration and fusion.
\newblock {\em IEEE Transactions on Pattern Analysis and Machine Intelligence}, 43(10):3333--3348, 2020.

\bibitem{shi2021unsharp}
Zenglin Shi, Yunlu Chen, Efstratios Gavves, Pascal Mettes, and Cees~GM Snoek.
\newblock Unsharp mask guided filtering.
\newblock {\em IEEE transactions on image processing}, 30:7472--7485, 2021.

\bibitem{xu2022model}
Shuang Xu, Jiangshe Zhang, Jialin Wang, Kai Sun, Chunxia Zhang, Junmin Liu, and Junying Hu.
\newblock A model-driven network for guided image denoising.
\newblock {\em Information Fusion}, 85:60--71, 2022.

\bibitem{sheng2022frequency}
Zehua Sheng, Xiongwei Liu, Si-Yuan Cao, Hui-Liang Shen, and Huaqi Zhang.
\newblock Frequency-domain deep guided image denoising.
\newblock {\em IEEE Transactions on Multimedia}, 25:6767--6781, 2022.

\bibitem{oh2023robust}
Geunwoo Oh, Jonghee Back, Jae-Pil Heo, and Bochang Moon.
\newblock Robust image denoising of no-flash images guided by consistent flash images.
\newblock In {\em Proceedings of the AAAI Conference on Artificial Intelligence}, volume~37, pages 1993--2001, 2023.

\bibitem{kingma2014adam}
Diederik~P Kingma and Jimmy Ba.
\newblock Adam: A method for stochastic optimization.
\newblock {\em arXiv preprint arXiv:1412.6980}, 2014.

\bibitem{huang2025wavelet}
Jie Huang, Rui Huang, Jinghao Xu, Siran Peng, Yule Duan, and Liang-Jian Deng.
\newblock Wavelet-assisted multi-frequency attention network for pansharpening.
\newblock In {\em Proceedings of the AAAI Conference on Artificial Intelligence}, volume~39, pages 3662--3670, 2025.

\bibitem{arienzo2022full}
Alberto Arienzo, Gemine Vivone, Andrea Garzelli, Luciano Alparone, and Jocelyn Chanussot.
\newblock Full-resolution quality assessment of pansharpening: Theoretical and hands-on approaches.
\newblock {\em IEEE Geoscience and Remote Sensing Magazine}, 10(3):168--201, 2022.

\end{thebibliography}

\newpage
\appendix
\section{Appendix}
This supplementary material provides additional technical and experimental details that support the main paper. It is organized as follows:

\begin{itemize}
    \item \textbf{Sec.\ref{Additional Proofs} Additional Proofs:} We provide detailed theoretical proofs of Theorem \ref{thm:1}, Lemma \ref{lemma1}, Lemma \ref{lemma3}, Lemma \ref{lemma2}, and Theorem \ref{theorem2} related to the SSO-PGA.
    \item \textbf{Sec.\ref{Limitations} Limitations:} We discuss the known limitation of our work and how to address it.
    \item \textbf{Sec.\ref{Broader Impact} Broader Impact:} We reflect on the potential applications and societal impact of our proposed method and framework.
    \item \textbf{Sec.\ref{Datasets} Datasets:} We provide an overview of the datasets employed in this work.
    \item \textbf{Sec.\ref{Implementation Details} Implementation Details:} We describe the compute resources, hyperparameters, and training strategies used in our experiments.
    \item \textbf{Sec.\ref{Experimental Results on Real-world Dataset} Experimental Results on Real-world Dataset:} We provide the experimental results on a full-resolution dataset to indicate the strong potential of our SSO-PGA for real-world applications.
    \item \textbf{Sec.\ref{Additional Vision Tasks} Additional Comparison with Traditional Proximal Gradient Algorithm:} We provide additional comparison with traditional proximal gradient algorithm to validate the advantages of our method.
    \item \textbf{Sec.\ref{Additional Ablation Study} Additional Ablation Study:} We provide additional ablation studies to validate the effectiveness of each component of our method.
    \item \textbf{Sec.\ref{Numerical Experiments} Additional Numerical Experiments:} We provide additional numerical experiment results to further validate the advantages of our method.
    \item \textbf{Sec.\ref{Additional Experimental Results} Additional Visual Experimental Results:} We include extended visual comparisons to further validate the effectiveness of our approach.
\end{itemize}

\subsection{Additional Proofs}
\label{Additional Proofs}
\textbf{Proof of Theorem \ref{thm:1}}
\begin{proof}
Using the identity \( \sigma(z)+\sigma(-z)=1 \), we have:
\begin{equation}
\begin{aligned}
{SSO}_{\alpha}\left(\nabla \mathcal{E}(\boldsymbol{y}_i^{t-1})\right)-1
&=2\left[\sigma(-\nabla \mathcal{E}(\boldsymbol{y}_i^{t-1})-\alpha) + \sigma(\alpha)-1\right]\\
&= 2\left[\sigma(-\nabla \mathcal{E}(\boldsymbol{y}_i^{t-1})-\alpha) - \sigma(-\alpha)\right].
\end{aligned}
\end{equation}
 
According to the Lagrange Mean Value Theorem, there exists \(\xi_i^{t} \) between \(-\alpha\) and \(-\nabla \mathcal{E}(\boldsymbol{y}_i^{t-1}) - \alpha\) such that
\begin{equation}
\sigma(-\nabla \mathcal{E}(\boldsymbol{y}_i^{t-1}) - \alpha) - \sigma(-\alpha) = (-\nabla \mathcal{E}(\boldsymbol{y}_i^{t-1}))\,\sigma'(\xi_i^{t}),
\end{equation}
where \( \sigma'(z) = \sigma(z)[1-\sigma(z)] \in (0, \tfrac14],\quad \forall z \in \mathbb{R} \).

Therefore,
\begin{equation}
{SSO}_{\alpha}\left(\nabla \mathcal{E}(\boldsymbol{y}_i^{t-1})\right)-1 = -2\nabla \mathcal{E}(\boldsymbol{y}_i^{t-1})\,\sigma'(\xi_i^{t}).
\end{equation}

Set \(\theta_i^{t} = 2\sigma'(\xi_i^{t})\). Then \({SSO}_{\alpha}\left(\nabla \mathcal{E}(\boldsymbol{y}_i^{t-1})\right) = 1 - \theta_i^{t}\nabla \mathcal{E}(\boldsymbol{y}_i^{t-1})\), and since \(\sigma'(z) \in (0,\tfrac14]\), it follows that \(\theta_i^{t} \in (0,\tfrac12]\). Thus, we have:
\begin{equation}
    \boldsymbol{y}_i^t = \boldsymbol{y}_i^{t-1} \cdot {SSO}_{\alpha}\left(\nabla \mathcal{E}(\boldsymbol{y}_i^{t-1})\right) = \boldsymbol{y}_i^{t-1}- \boldsymbol{y}_i^{t-1}\theta_i^{t}\nabla \mathcal{E}(\boldsymbol{y}_i^{t-1}).
\end{equation}

Set \(\rho_i = \boldsymbol{y}_i^{t-1}\theta_i^{t}\), proof complete.
\end{proof}

\textbf{Proof of Lemma \ref{lemma1}}

\begin{proof}
Recall that the sliding sigmoid operator is defined as:
\begin{equation}
SSO_\alpha(z) = 2\sigma(-z - \alpha) + 2\sigma(\alpha) - 1, \quad \text{where} \quad \sigma(u) = \frac{1}{1 + e^{-u}}.
\end{equation}

Since $SSO_\alpha(0) = 1$, by the Lagrange Mean Value Theorem, for some \( \xi \in (0, z) \) (or \( (z, 0) \)), we have:
\begin{equation}
SSO_\alpha(z) - 1 = SSO_\alpha'(\xi) \cdot z.
\end{equation}

Now compute the derivative:
\begin{equation}
SSO_\alpha'(u) = \frac{d}{du}\left[2\sigma(-u - \alpha)\right] = -2\sigma(-u - \alpha)(1 - \sigma(-u - \alpha)).
\end{equation}

The maximum of \( \sigma(v)(1 - \sigma(v)) \) over \( v \in \mathbb{R} \) is \( \frac{1}{4} \), hence:
\begin{equation}
|SSO_\alpha'(u)| \le \frac{1}{2} \le \frac{1+{\alpha}}{2}= \eta(\alpha).
\end{equation}

Thus:
\begin{equation}
|SSO_\alpha(z) - 1| \le |SSO_\alpha'(\xi)| \cdot |z| \le \eta(\alpha) |z|.
\end{equation}
\end{proof}

\textbf{Proof of Lemma \ref{lemma3}}
\begin{proof}
Consider the scalar function 
\( 
    \varphi(t)\;=\;\mathcal{E}(\boldsymbol{y}+t\boldsymbol{d}),
    \quad t\in[0,1].
\)
We have:
\begin{equation}
    \mathcal{E}(\boldsymbol{y}+\boldsymbol{d})-\mathcal{E}(\boldsymbol{y})
    =\varphi(1)-\varphi(0)
    =\int_{0}^{1}\varphi'(t)\,dt
    =\int_{0}^{1}
      \bigl\langle\nabla\mathcal{E}(\boldsymbol{y}+t\boldsymbol{d}),\boldsymbol{d}\bigr\rangle
      dt.
\end{equation}

Add and subtract $\nabla\mathcal{E}(\boldsymbol{y})$ inside the inner product and apply Cauchy–Schwarz:
\begin{equation}
\begin{aligned}
    \mathcal{E}(\boldsymbol{y}+\boldsymbol{d})-\mathcal{E}(\boldsymbol{y})
    &=\int_{0}^{1}
      \bigl\langle\nabla\mathcal{E}(\boldsymbol{y}),\boldsymbol{d}\bigr\rangle dt
      +\int_{0}^{1}
      \bigl\langle
        \nabla\mathcal{E}(\boldsymbol{y}+t\boldsymbol{d})
        -\nabla\mathcal{E}(\boldsymbol{y}),
        \boldsymbol{d}
      \bigr\rangle dt
\\
    &=\langle\nabla\mathcal{E}(\boldsymbol{y}),\boldsymbol{d}\rangle
      +\int_{0}^{1}
        \bigl\langle
          \nabla\mathcal{E}(\boldsymbol{y}+t\boldsymbol{d})
          -\nabla\mathcal{E}(\boldsymbol{y}),
          \boldsymbol{d}
        \bigr\rangle dt
\\
    &\le
      \langle\nabla\mathcal{E}(\boldsymbol{y}),\boldsymbol{d}\rangle
      +\int_{0}^{1}
        \|\nabla\mathcal{E}(\boldsymbol{y}+t\boldsymbol{d})
          -\nabla\mathcal{E}(\boldsymbol{y})\|_2\;\|\boldsymbol{d}\|_2 \, dt
\\
    &\le
      \langle\nabla\mathcal{E}(\boldsymbol{y}),\boldsymbol{d}\rangle
      +\int_{0}^{1} L\, t\,\|\boldsymbol{d}\|_2^{2}\,dt
      \quad\text{(by $L$--Lipschitzness)}
\\
    &=
      \langle\nabla\mathcal{E}(\boldsymbol{y}),\boldsymbol{d}\rangle
      +\frac{L}{2}\,\|\boldsymbol{d}\|_2^{2}.
\end{aligned}
\end{equation}

Thus, proof complete.
\end{proof}

\textbf{Proof of Lemma \ref{lemma2}}
\begin{proof}
The gradient of the objective is:
\begin{equation}
\nabla \mathcal{E}(\boldsymbol{y}) = 2\boldsymbol{H}^\top (\boldsymbol{H}\boldsymbol{y} - \boldsymbol{x}).
\end{equation}

So for any \( \boldsymbol{y}, \boldsymbol{z} \), we have:
\begin{equation}
\begin{aligned}
\| \nabla \mathcal{E}(\boldsymbol{y}) - \nabla \mathcal{E}(\boldsymbol{z}) \|_2
&= 2 \| \boldsymbol{H}^\top \boldsymbol{H} (\boldsymbol{y} - \boldsymbol{z}) \|_2 \\
&\le 2 \| \boldsymbol{H}^\top \boldsymbol{H} \|_2 \cdot \| \boldsymbol{y} - \boldsymbol{z} \|_2 \\
&= 2 \| \boldsymbol{H} \|_2^2 \cdot \| \boldsymbol{y} - \boldsymbol{z} \|_2.
\end{aligned}
\end{equation}

Thus,
\begin{equation}
  \|\nabla\mathcal E(\boldsymbol{y})-\nabla\mathcal E(\boldsymbol{z})\|_2\le L\|\boldsymbol{y}-\boldsymbol{z}\|_2,\quad L=2\|\boldsymbol{H}\|_2^{2}.
\end{equation}
\end{proof}

\textbf{Proof of Theorem~\ref{theorem2}}
\begin{proof}
Fix \(t\), and denote \(\boldsymbol{y}=\boldsymbol{y}^{t-1}\), \(\boldsymbol{y}^{+}=\boldsymbol{y}^{t}\), \(\boldsymbol{g}=\nabla\mathcal E(\boldsymbol{y})\) and \(\boldsymbol{s}=SSO_{\alpha}(\boldsymbol{g})-\boldsymbol{1}\) for simplicity.. From Eq.~(\ref{eq: proof1}), we have \(\boldsymbol{y}^{+}=\boldsymbol{y}+\boldsymbol{d}\) with \(\boldsymbol{d}=\boldsymbol{y}\odot\boldsymbol{s}\). 
Then, we have:
\begin{equation}
  \langle \boldsymbol{g},\boldsymbol{d}\rangle = -\sum_{i}|d_{i}||g_{i}|,
\end{equation}
\textbf{Lemma~\ref{lemma1}} with \(z=g_{i}\) yields \(|s_{i}|\le\eta(\alpha)|g_{i}|\). Hence:
\begin{equation}
\|\boldsymbol{d}\|_2^{2}=\sum_{i}|d_{i}||s_{i}|y_{i}\le\eta(\alpha)\sum_{i}|d_{i}||g_{i}|y_{i}.
\end{equation}
From \textbf{Lemma~\ref{lemma2}}, \(\alpha\le 2/(\kappa\|\boldsymbol{H}\|_2^{2})-1=4/(\kappa L)-1\), we have \(\eta(\alpha)=(\alpha+1)/2\le2/(\kappa L)\). Combining this with the bound on \(\|\boldsymbol{d}\|_2^2\) gives:
\begin{equation}
  \frac{L}{2}\|\boldsymbol{d}\|_2^{2}\le\sum_{i}|d_{i}||g_{i}| = -\langle \boldsymbol{g},\boldsymbol{d}\rangle.
\end{equation}

Inserting the bounds into \textbf{Lemma~\ref{lemma3}}:
\begin{align}
  \mathcal E(\boldsymbol{y}^{+})-\mathcal E(\boldsymbol{y})
  &\le\langle \boldsymbol{g},\boldsymbol{d}\rangle+\frac{L}{2}\|\boldsymbol{d}\|_2^{2}\\
  &\le\langle \boldsymbol{g},\boldsymbol{d}\rangle-\langle \boldsymbol{g},\boldsymbol{d}\rangle
  =0.
\end{align}

Thus, \(\mathcal E(\boldsymbol{y}^{t})\le\mathcal E(\boldsymbol{y}^{t-1})\) for every \(t\ge1\).
\end{proof}

\subsection{Limitations}
\label{Limitations}
\begin{figure}[h!]
\centering
 \includegraphics[width = 0.99\textwidth]{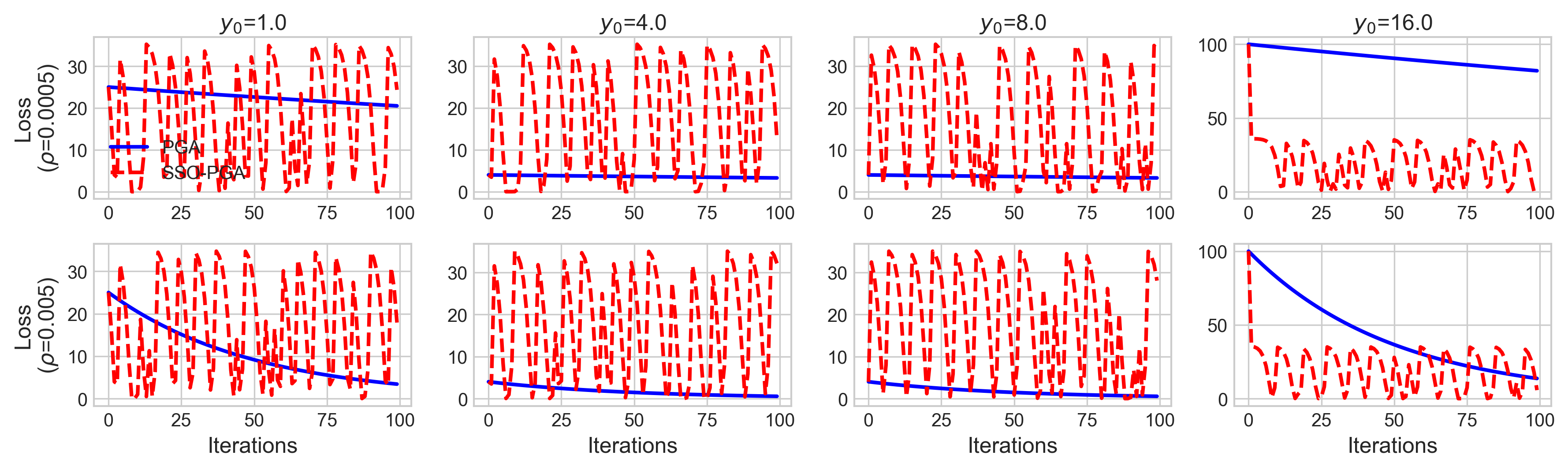}
\caption{Comparison of numerical simulation results for SSO-PGA and PGA on Problem I when the true solution is large.}
\label{fig:lim1}
\end{figure}

\begin{figure}[h!]
\centering
 \includegraphics[width = 0.99\textwidth]{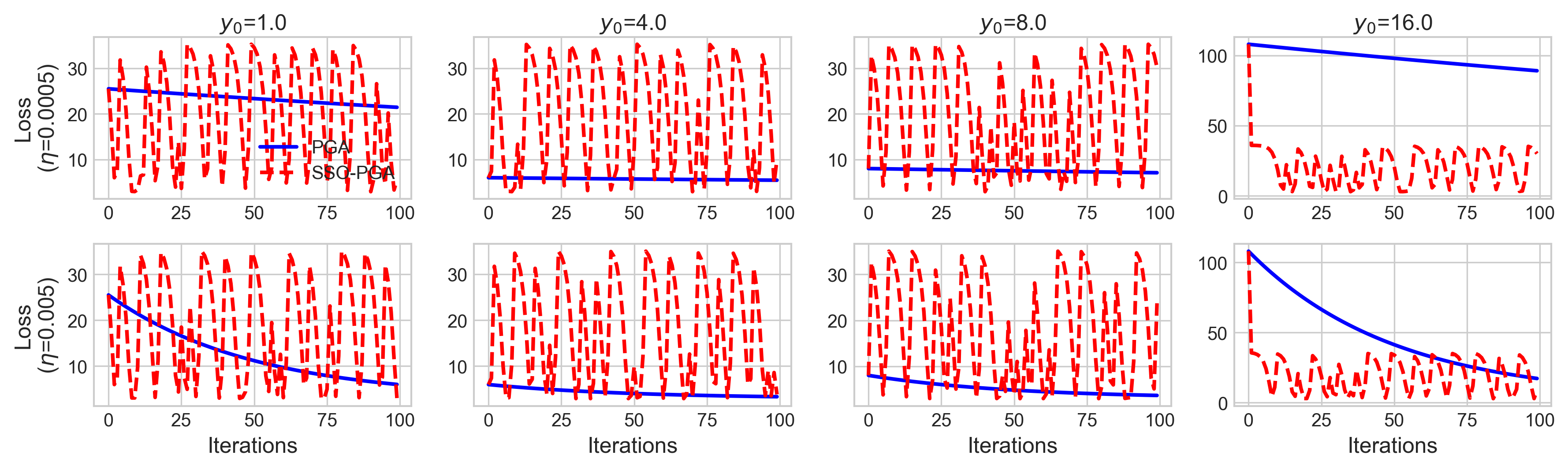}
\caption{Comparison of numerical simulation results for SSO-PGA and PGA on Problem II when the true solution is large.}
\label{fig:lim3}
\end{figure}

A limitation of our study is that SSO-PGA performs well when the solution to the optimization problem lies between 0 and 1, but exhibits oscillatory, non-convergent behavior when the true solution is large. For example, when we set the optimal solution to 6, as shown in Fig.~\ref{fig:lim1} and Fig.~\ref{fig:lim3}, this issue becomes apparent.

This specific instability is circumvented when SSO-PGA is integrated with a deep network. This is because, in deep learning, it's standard practice to normalize network inputs and outputs to the [0, 1] range. The final results are then obtained through inverse normalization. This preprocessing step naturally prevents the instability observed with large solution values.

\begin{figure}[h!]
\centering
 \includegraphics[width = 0.99\textwidth]{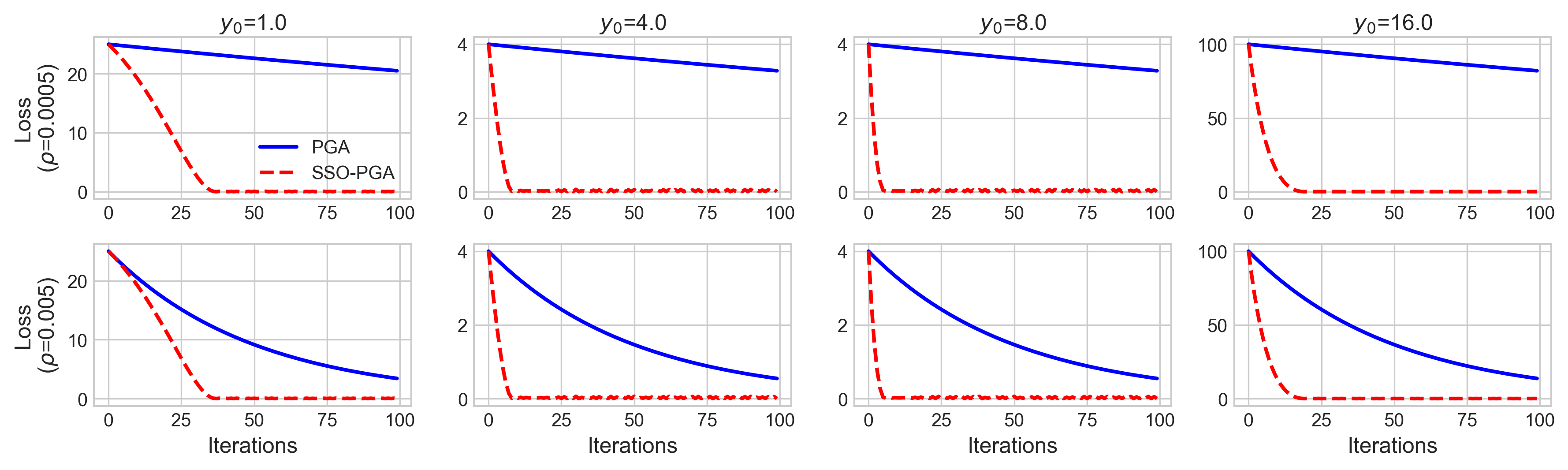}
\caption{Comparison of numerical simulation results for SSO-PGA (with gradient clipping) and PGA on Problem I when the true solution is large.}
\label{fig:lim2}
\end{figure}

\begin{figure}[h!]
\centering
 \includegraphics[width = 0.99\textwidth]{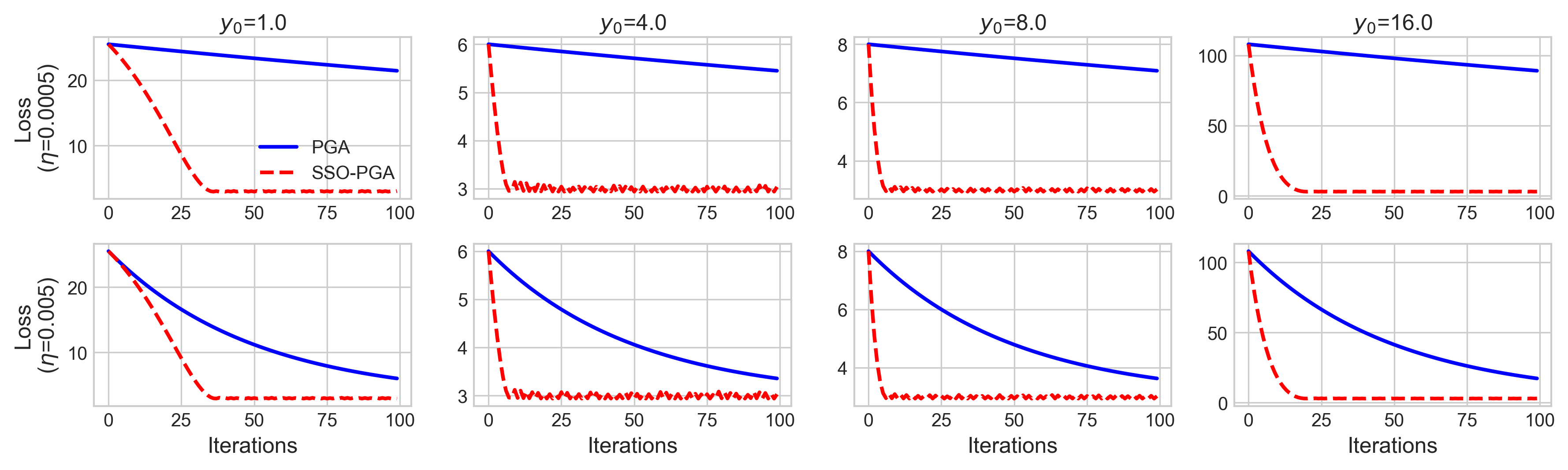}
\caption{Comparison of numerical simulation results for SSO-PGA (with gradient clipping) and PGA on Problem II when the true solution is large.}
\label{fig:lim4}
\end{figure}

Furthermore, we've identified that this oscillatory behavior is caused by excessively large gradients. We propose a straightforward solution to mitigate this problem during the optimization process: gradient clipping. For instance, by clipping the gradients of SSO-PGA to a range of [-0.1, 0.1], as shown in Fig.~\ref{fig:lim2} and Fig.~\ref{fig:lim4}, SSO-PGA still demonstrates a faster convergence rate compared to PGA.

\subsection{Broader Impact}
\label{Broader Impact}
The proposed SSO-PGA framework offers a robust and interpretable optimization strategy that extends beyond the task of multispectral image fusion and flash guided non-flash image denoising. The SSO-PGA framework is built upon a gradient-based update mechanism that naturally enforces non-negativity, making it easily adaptable to various inverse problems in computer vision and image reconstruction tasks. These include, but are not limited to, image deblurring, denoising, super-resolution, compressive sensing reconstruction, and medical image enhancement. The deep unfolding nature of SSO-PGA not only enables convergence-guaranteed iterative learning but also offers structural transparency, which is particularly desirable in safety-critical applications like healthcare and autonomous navigation. The strong empirical performance and theoretical convergence guarantee of SSO-PGA make it a promising foundation for future research on interpretable and robust optimization in deep learning systems.

\subsection{Datasets}
\label{Datasets}
In our experiments on multispectral image fusion, we utilized remote sensing image datasets from the PanCollection repository \cite{deng2022machine}, encompassing three satellite sources: WorldView-3 (WV3), QuickBird (QB), and GaoFen-2 (GF2). Each dataset is divided into training and testing subsets. A detailed summary of the sample counts and image dimensions under both reduced- and full-resolution settings is provided in Table~\ref{tab:dataset_summary}.
\begin{table}[htbp]
\centering
\caption{Summary of WorldView-3 (WV3), QuickBird (QB), and GaoFen-2 (GF2) datasets.}
\renewcommand{\arraystretch}{1.2}
\setlength{\tabcolsep}{8pt}
\begin{tabular}{lcc}
\toprule
Dataset & Samples  & Image Size (PAN / LRMS / GT) \\
\midrule
\multicolumn{3}{c}{\textit{Reduced-Resolution}} \\
\midrule
WV3 & 10,000 (train) / 20 (test)  & 64$\times$64 / 16$\times$16$\times$8 / 64$\times$64$\times$8 \\
QB  & 17,000 (train) / 20 (test)  & 64$\times$64 / 16$\times$16$\times$4 / 64$\times$64$\times$4 \\
GF2 & 20,000 (train) / 20 (test)  & 64$\times$64 / 16$\times$16$\times$4 / 64$\times$64$\times$4 \\
\midrule
\multicolumn{3}{c}{\textit{Full-Resolution}} \\
\midrule
WV3 & 20 (test)  & 512$\times$512 / 128$\times$128$\times$8 / None \\
QB  & 20 (test)  & 512$\times$512 / 128$\times$128$\times$4 / None \\
GF2 & 20 (test)  & 512$\times$512 / 128$\times$128$\times$4 / None \\
\bottomrule
\end{tabular}
\label{tab:dataset_summary}
\end{table}

In our experiments on flash guided non-flash image denoising, we utilized two common datasets: the Flash and Ambient Illuminations Dataset (FAID) \cite{aksoy2018dataset} and the Multi-Illumination Dataset (MID) \cite{murmann2019dataset}. Each dataset is divided into training and testing subsets. A detailed summary of the sample counts and image dimensions is provided in Table~\ref{tab:dataset_summary2}.
\begin{table}[htbp]
\centering
\caption{Summary of the Flash and Ambient Illuminations Dataset (FAID) \cite{aksoy2018dataset} and the Multi-Illumination Dataset (MID) \cite{murmann2019dataset}.}
\renewcommand{\arraystretch}{1.2}
\setlength{\tabcolsep}{8pt}
\begin{tabular}{lcc}
\toprule
Dataset & Samples  & Image Size\\
\midrule
FAID & 404 (train) / 12 (test)  & 900$\times$600 $\times$ 3 \\
MID  & 983 (train) / 30 (test)  & 1500$\times$1000 $\times$ 3  \\
\bottomrule
\end{tabular}
\label{tab:dataset_summary2}
\end{table}

\subsection{Implementation Details}
\label{Implementation Details}
All training procedures are conducted on a high-performance computing server equipped with $8$ NVIDIA RTX $4090$ GPUs. Our training pipeline is implemented in Python $3.8.20$ with PyTorch $2.4.1 + \text{cu}121$, leveraging CUDA $12.1$ for efficient GPU acceleration. 

For multispectral image fusion, we employ the Adam optimizer \cite{kingma2014adam} with an initial learning rate of $1 \times 10^{-3}$ and a weight decay of $1 \times 10^{-8}$, and the learning rate is halved every 100 epochs. The model is trained for 300 epochs. During training, we apply dropout regularization with rates of 0.1 on the WV3 and QB datasets, and 0.25 on the GF2 dataset. To ensure high-quality reconstruction, we adopt a batch size of 32 throughout the training process. The entire model contains approximately 1.07 million trainable parameters and requires around 15.20 GiB of GPU memory. We compare our method with several state-of-the-art methods, including 3 traditional algorithms: MTF-GLP-FS \cite{vivone2018full}, BDSD-PC \cite{vivone2019robust}, and TV \cite{palsson2013new}, and 11 deep learning/unfolding-based models: PNN \cite{masi2016pansharpening}, PanNet \cite{yang2017pannet}, DiCNN \cite{he2019pansharpening}, FusionNet \cite{deng2020detail}, MDCUN \cite{yang2022memory}, LAGNet \cite{jin2022lagconv}, LGPNet \cite{zhao2023lgpconv}, U2Net \cite{peng2023u2net}, CANNet \cite{duan2024content}, PanMamba \cite{he2025pan}, and ADWM \cite{huang2025general}.

For flash guided non-flash image denoising, we employ the Adam optimizer \cite{kingma2014adam} with an initial learning rate of $1 \times 10^{-3}$ and a weight decay of $1 \times 10^{-8}$, and the learning rate is halved every 300 epochs. The model is trained for 2000 epochs. To ensure high-quality reconstruction, we adopt a batch size of 16 and a patch size of 128 × 128 throughout the training process. The entire model contains approximately 2.90 million trainable parameters and requires around 39.91 GiB of GPU memory. We compared our results against the following representative methods: DnCNN \cite{zhang2017beyond}, 
DJFR \cite{li2019joint}, CUNet \cite{deng2020deep}, UMGF \cite{shi2021unsharp}, MN \cite{xu2022model}, FGDNet \cite{sheng2022frequency}, RIDFhF \cite{oh2023robust}, and DeepM$^2$CDL \cite{10323520}.

\subsection{Experimental Results on Real-world Dataset}
\label{Experimental Results on Real-world Dataset}
Following \cite{huang2025wavelet}, for full-resolution data, we apply D$_s$, D$_\lambda$, and HQNR \cite{arienzo2022full} as the evaluation metric, which collectively provide a comprehensive measure of image fusion quality. We evaluate SSO-PGA on the full-resolution WV3 dataset, where it demonstrates significant advantages in Tab.~\ref{tab:fr_metrics}. This outstanding performance not only validates the effectiveness of our method but also underscores its robustness and profound potential for real-world applications requiring high-fidelity image fusion.
\begin{table}[htbp]
\centering
\caption{Quantitative comparison on WV3 dataset with 20 full-resolution samples. The best results are in \textbf{bold} and the second-best values are \underline{underlined}.}
\renewcommand{\arraystretch}{1.2}
\setlength{\tabcolsep}{3.6pt}
\resizebox{\textwidth}{!}{%
\begin{tabular}{c|cccc}
\toprule
Methods    & BDSD-PC \cite{vivone2019robust} & TV \cite{palsson2013new} &  PNN \cite{masi2016pansharpening} & PanNet \cite{yang2017pannet}\\
\midrule
$D_\lambda \downarrow$   & 0.063 & 0.023 & 0.021 & \textbf{0.017}\\
$D_s \downarrow$         & 0.073 & 0.039 & 0.043 & 0.047\\
HQNR $\uparrow$          & 0.870 & 0.938 & 0.937 & 0.937\\
\midrule
Methods    & DiCNN \cite{he2019pansharpening}  & LAGNet \cite{jin2022lagconv} & LGPNet \cite{zhao2023lgpconv}  & U2Net \cite{peng2023u2net} \\
\midrule
$D_\lambda \downarrow$   & 0.036  & 0.037 & 0.022 & 0.020 \\
$D_s \downarrow$         & 0.046 & 0.042 & 0.039 & \underline{0.028} \\
HQNR $\uparrow$          & 0.920 & 0.923 & 0.940 & \underline{0.952} \\
\midrule
Methods    & CANNet \cite{duan2024content} & PanMamba \cite{he2025pan} & ADWM \cite{huang2025general} & SSO-PGA (ours)  \\
\midrule
$D_\lambda \downarrow$    & 0.020 & \underline{0.018} & 0.024 & 0.022 \\
$D_s \downarrow$          & 0.030 & 0.053 & 0.029 & \textbf{0.026} \\
HQNR $\uparrow$           & 0.951 & 0.930 & 0.948 & \textbf{0.953} \\
\bottomrule
\end{tabular}
}
\label{tab:fr_metrics}
\end{table}

\subsection{Additional Comparison with Traditional Proximal Gradient Algorithm}
\label{Additional Vision Tasks}
In this subsection, we provide a supplementary perturbation analysis for both PGA and SSO-PGA. Additionally, we present further experimental results for SSO-PGA at different iteration counts, as detailed in Tab.~\ref{tab:itr}.

\begin{figure}[H]
\centering
 \includegraphics[width = 0.6\textwidth]{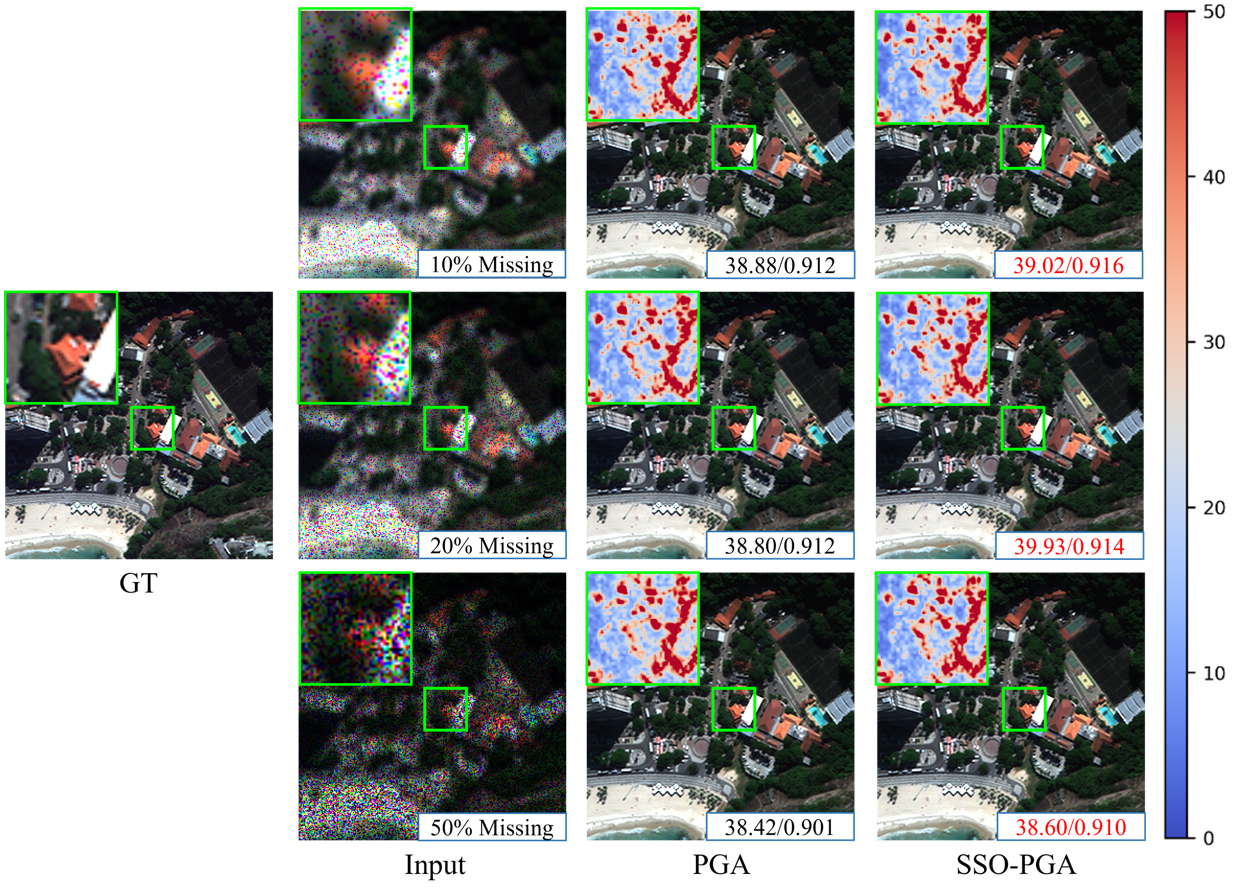}
\caption{Visual comparison along with the corresponding PSNR and Q8 values of SSO-PGA and PGA on the WV3 dataset under varying missing ratios.}
\label{fig:missing}
\end{figure}

\textbf{Perturbation Analysis.}
Fig.~\ref{fig:missing} and Tab.~\ref{tab:miss} present the comparison between SSO-PGA and PGA under varying levels of missing MS input ($10\%$, $20\%$, and $50\%$) on the WV3 dataset. Across all perturbation levels, SSO-PGA consistently yields superior visual reconstruction and achieves higher PSNR and Q8 scores compared to PGA. Especially under a high missing rate ($50\%$), the Q8 value of PGA drops to only 0.901, while SSO-PGA still maintains a result of 0.910. This demonstrates the strong robustness of the proposed SSO-PGA method in handling degraded and incomplete inputs.

In conclusion, comparing SSO-PGA with the PGA baseline, the results in Tab.~\ref{tab:itr} validate that SSO-PGA achieves faster and more stable convergence, while the perturbation experiments in Tab.~\ref{tab:miss} confirm its robustness under various missing ratios.

\begin{table}[ht]
\centering
\caption{Quantitative comparison of SSO-PGA and PGA  on the WV3 reduced-resolution dataset over different iterations. The better results are in \textbf{bold}.}
\renewcommand{\arraystretch}{1.2}
\setlength{\tabcolsep}{3.6pt}
\resizebox{0.8\textwidth}{!}{%
\begin{tabular}{l|cccc|cccc}
\toprule
\multirow{3}{*}{Iteration 1} & \multicolumn{4}{c|}{PGA} & \multicolumn{4}{c}{SSO-PGA} \\
 & PSNR↑ & SAM↓ & ERGAS↓ & Q8↑ & PSNR↑ & SAM↓ & ERGAS↓ & Q8↑  \\
    & 38.882	&2.961	&\textbf{2.193}	&0.916 & \textbf{38.900}	&\textbf{2.960}	&2.200	&\textbf{0.917} \\
    \midrule
\multirow{3}{*}{Iteration 2} & \multicolumn{4}{c|}{PGA} & \multicolumn{4}{c}{SSO-PGA} \\
 & PSNR↑ & SAM↓ & ERGAS↓ & Q8↑ & PSNR↑ & SAM↓ & ERGAS↓ & Q8↑ \\
   & 39.027	&2.944	&2.160	&0.916 & \textbf{39.131}&\textbf{2.892}	&\textbf{2.138}	&\textbf{0.918}	 \\
   \midrule
\multirow{3}{*}{Iteration 3} & \multicolumn{4}{c|}{PGA} & \multicolumn{4}{c}{SSO-PGA} \\
 & PSNR↑ & SAM↓ & ERGAS↓ & Q8↑ & PSNR↑ & SAM↓ & ERGAS↓ & Q8↑ \\
   & 39.114	&2.913	&2.142	&0.919 & \textbf{39.256}	&\textbf{2.855}	&\textbf{2.104}	&\textbf{0.920}	 \\
   \midrule
\multirow{3}{*}{Iteration 4} & \multicolumn{4}{c|}{PGA} & \multicolumn{4}{c}{SSO-PGA} \\
 & PSNR↑ & SAM↓ & ERGAS↓ & Q8↑ & PSNR↑ & SAM↓ & ERGAS↓ & Q8↑ \\
   & 39.145	&2.925	&2.129	&0.918 & \textbf{39.358}	&\textbf{2.823}	&\textbf{2.078}	&\textbf{0.921} \\
   \midrule
\multirow{3}{*}{Iteration 5} & \multicolumn{4}{c|}{PGA} & \multicolumn{4}{c}{SSO-PGA} \\
 & PSNR↑ & SAM↓ & ERGAS↓ & Q8↑ & PSNR↑ & SAM↓ & ERGAS↓ & Q8↑ \\
   & 39.125	&2.916	&2.139	&0.918	&\textbf{39.374}	&\textbf{2.818}	&\textbf{2.072}	&\textbf{0.921}	 \\
\bottomrule
\end{tabular}
}
\label{tab:itr}
\end{table}

\begin{table}[ht]
\centering
\caption{Quantitative comparison of SSO-PGA and PGA  on the WV3 reduced-resolution dataset under varying missing ratios. The better results are in \textbf{bold}.}
\renewcommand{\arraystretch}{1.2}
\setlength{\tabcolsep}{3.6pt}
\resizebox{0.8\textwidth}{!}{%
\begin{tabular}{l|cccc|cccc}
\toprule
\multirow{3}{*}{Missing 10\%} & \multicolumn{4}{c|}{PGA} & \multicolumn{4}{c}{SSO-PGA} \\
 & PSNR↑ & SAM↓ & ERGAS↓ & Q8↑ & PSNR↑ & SAM↓ & ERGAS↓ & Q8↑  \\
    & 38.875	&3.039	&2.196	&0.912 & \textbf{39.016} & \textbf{2.931} & \textbf{2.157} & \textbf{0.916} \\
    \midrule
\multirow{3}{*}{Missing 20\%} & \multicolumn{4}{c|}{PGA} & \multicolumn{4}{c}{SSO-PGA} \\
 & PSNR↑ & SAM↓ & ERGAS↓ & Q8↑ & PSNR↑ & SAM↓ & ERGAS↓ & Q8↑ \\
   &38.804	&3.056	&2.210	&0.912  & \textbf{38.928} & \textbf{2.972} & \textbf{2.184} & \textbf{0.914}	 \\
   \midrule
\multirow{3}{*}{Missing 30\%} & \multicolumn{4}{c|}{PGA} & \multicolumn{4}{c}{SSO-PGA} \\
 & PSNR↑ & SAM↓ & ERGAS↓ & Q8↑ & PSNR↑ & SAM↓ & ERGAS↓ & Q8↑ \\
   & 38.420	&3.415	&2.301	&0.901 & \textbf{38.599} & \textbf{3.079} & \textbf{2.270} & \textbf{0.910}	 \\
\bottomrule
\end{tabular}
}
\label{tab:miss}
\end{table}

\subsection{Additional Ablation Study}
\label{Additional Ablation Study}
\textbf{Different Sliding Parameter Settings. }
Besides the parameter learning method described in the deep network architecture, there are two other ways to set the sliding parameter: manually fixed value and automated learning via a simple neural network (with a Convolution layer, a Sigmoid activation, another Convolution layer, and finally a Softplus activation). We've conducted additional experiments to compare these two approaches (Tab.~\ref{tab:ab1}), where the SSO-PGA-1 and SSO-PGA-0.1 are our method with fixed $\alpha$ values (1/0.1), and SSO-PGA-Auto is the automated way. From the table, we can observe that the performance of the fixed-value sliding parameter and the automated approach is slightly lower than that of our method in the paper.
\begin{table}[h]
\centering
\caption{Comparison of Different Sliding Parameter Settings.}
\label{tab:ab1}
\begin{tabular}{lcccc}
\toprule
 & PSNR $\uparrow$ & SAM $\downarrow$ & ERGAS $\downarrow$ & Q2N $\downarrow$ \\
\midrule
SSO-PGA-Auto & 39.280 & 2.841 & 2.099 & \textbf{0.921} \\
SSO-PGA-1 & 39.287 & 2.824 & 2.092 & \textbf{0.921} \\
SSO-PGA-0.1 & 39.147 & 2.884 & 2.115 & 0.920 \\
SSO-PGA & \textbf{39.358} & \textbf{2.823} & \textbf{2.078} & \textbf{0.921} \\
\bottomrule
\end{tabular}
\end{table}

\textbf{Comparison with Traditional Projected Operator. }
To compare with traditional post-projection methods, we attempted to enforce non-negativity by applying activation functions (ReLU and Softplus) as projection operations after the gradient descent step in traditional PGA. The experimental results are shown in Tab.~\ref{tab:ab2}. However, both approaches performed even worse than PGA. The reason for this is that while these projection methods enforce non-negativity, they unfortunately lose information from negative values and alter the original gradient information during the process. In contrast, SSO-PGA guarantees non-negativity through a direct mapping while fully preserving the gradient information.
\begin{table}[h]
\centering
\caption{Comparison with Traditional Projected Gradient Descent Methods.}
\label{tab:ab2}
\begin{tabular}{lcccc}
\toprule
 & PSNR $\uparrow$ & SAM $\downarrow$ & ERGAS $\downarrow$ & Q2N $\downarrow$ \\
\midrule
ReLU-PGA & 36.600 & 3.557 & 2.825 & 0.900 \\
Softplus-PGA & 38.957 & 2.926 & 2.167 & 0.916 \\
PGA & 39.145 & 2.925 & 2.129 & 0.918 \\
SSO-PGA & \textbf{39.358} & \textbf{2.823} & \textbf{2.078} & \textbf{0.921} \\
\bottomrule
\end{tabular}
\end{table}

\subsection{Additional Numerical Experiments}
\label{Numerical Experiments}

\begin{figure}[H]
\centering
 \includegraphics[width = 0.85\textwidth]{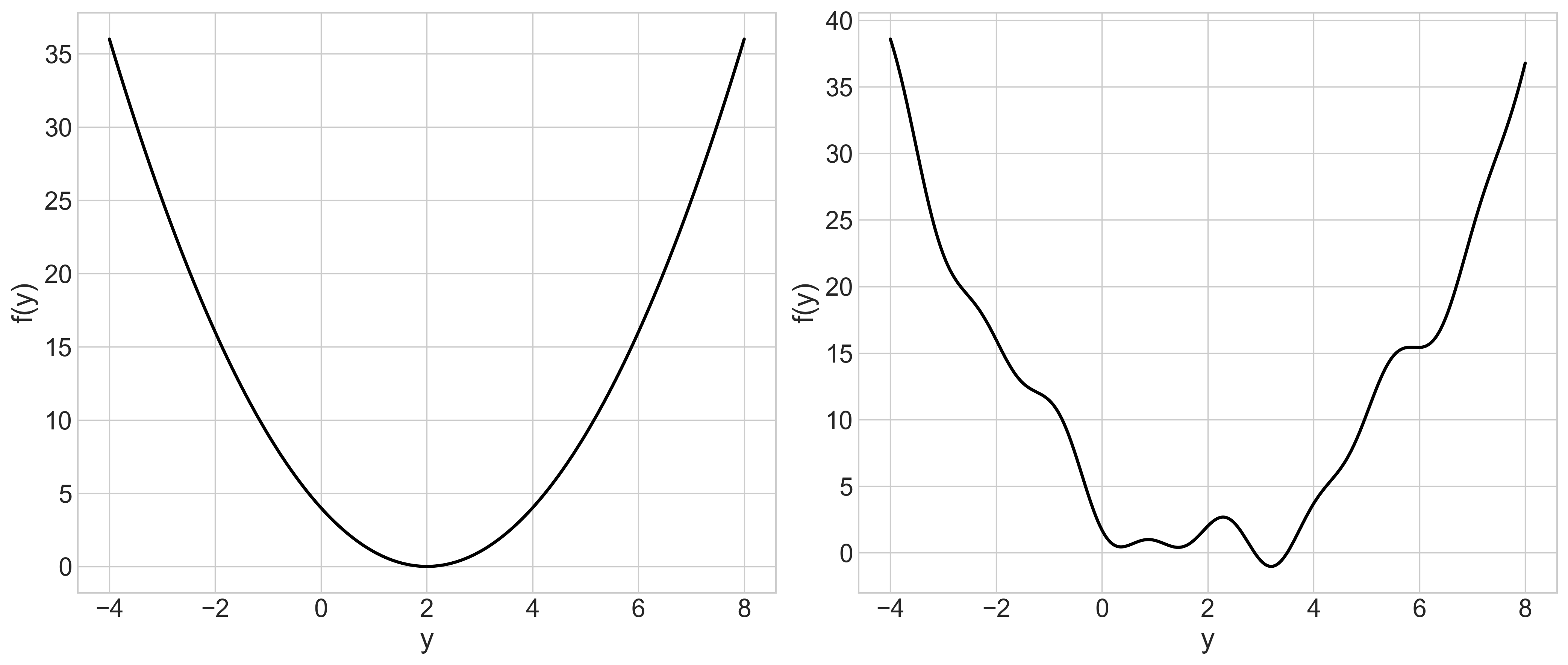}
\caption{Landscapes for Problem I (left) and Problem I+ (right).}
\label{fig:function}
\end{figure}

In this subsection, we provide additional numerical simulation experiments. Specifically, in addition to the two problems from Eq.~(\ref{eq: numerical simulation experiments}), we include two non-convex problems, denoted as Problem I+ and Problem II+:
\begin{equation}
\begin{aligned}
&\min_{y} ({y}-0.5)^2 +\sin(4(x-0.5)) + \cos(2(x-0.5)),\quad &&(\text{Problem I+}),\\
    &\min_{y} ({y}-0.5)^2 +\sin(4(x-0.5)) + \cos(2(x-0.5))+\frac{1}{2}|y|,\quad &&(\text{Problem II+}).
\label{eq: numerical simulation experiments2}
\end{aligned}
\end{equation}

\begin{figure}
\centering
 \includegraphics[width = \textwidth]{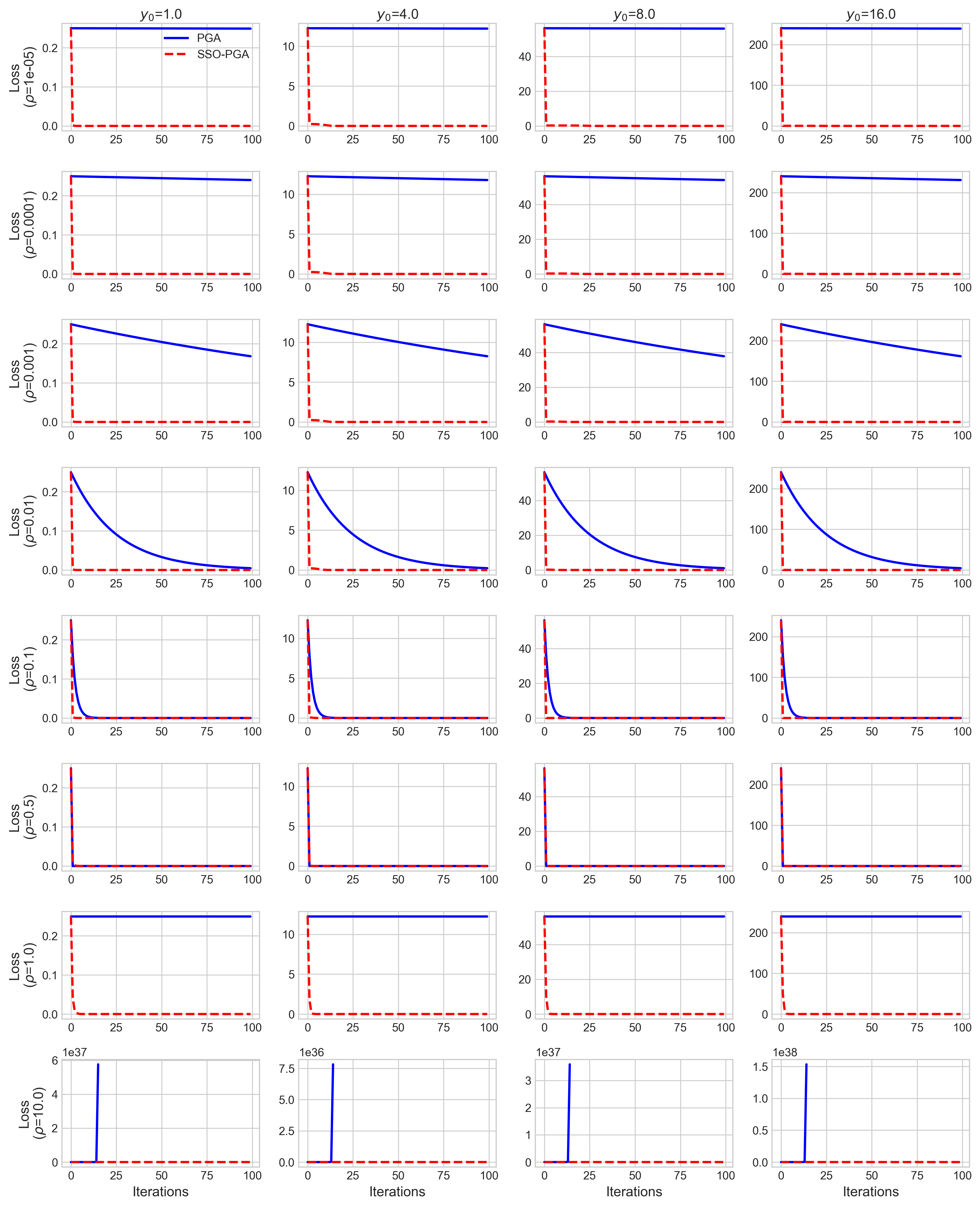}
\caption{Additional comparison of numerical simulation results for SSO-PGA and PGA on Problem I.}
\label{fig:Numerical_Experiments1}
\end{figure}

\begin{figure}[H]
\centering
 \includegraphics[width = \textwidth]{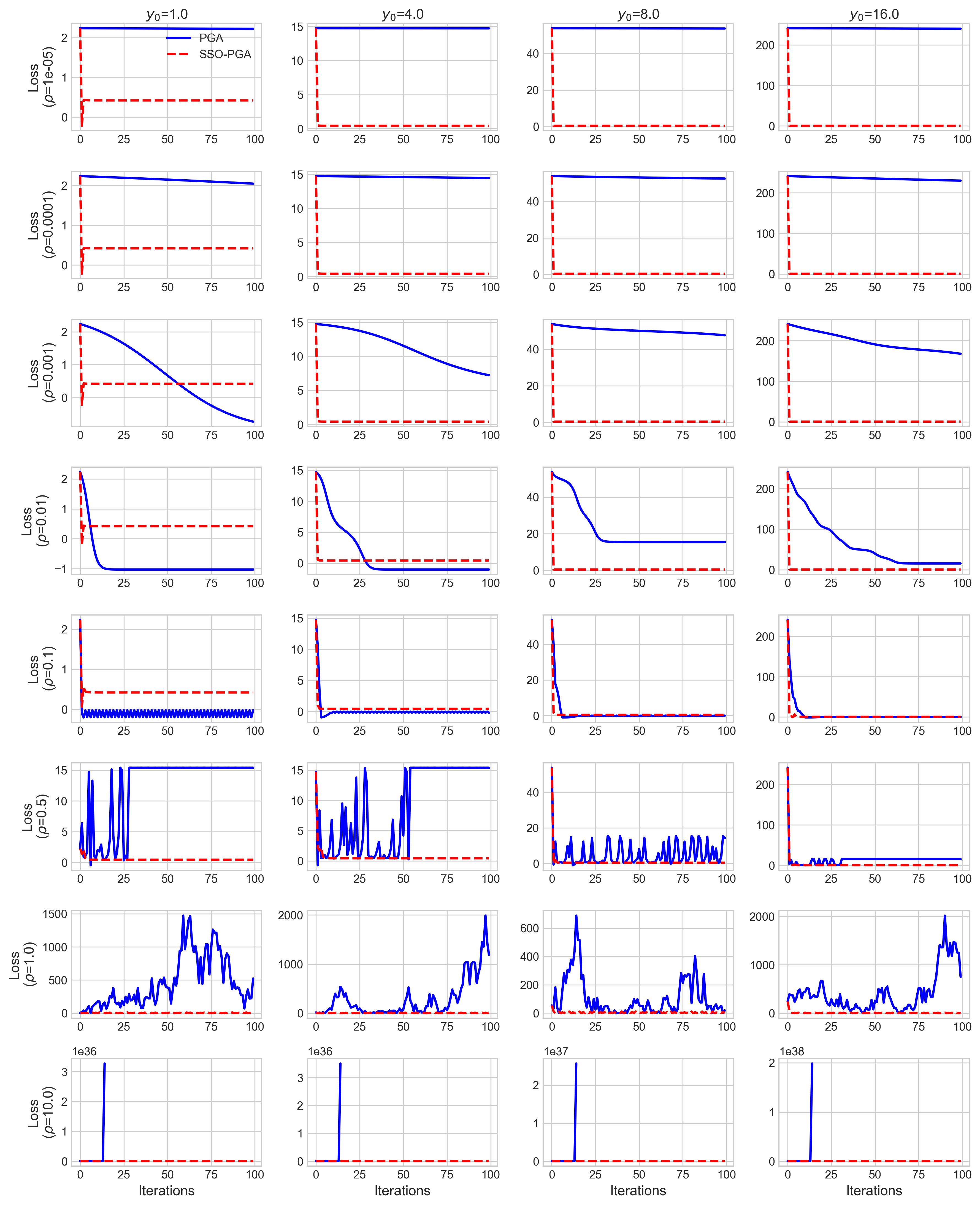}
\caption{Additional comparison of numerical simulation results for SSO-PGA and PGA on Problem I+.}
\label{fig:Numerical_Experiments2}
\end{figure}

\begin{figure}[H]
\centering
 \includegraphics[width = \textwidth]{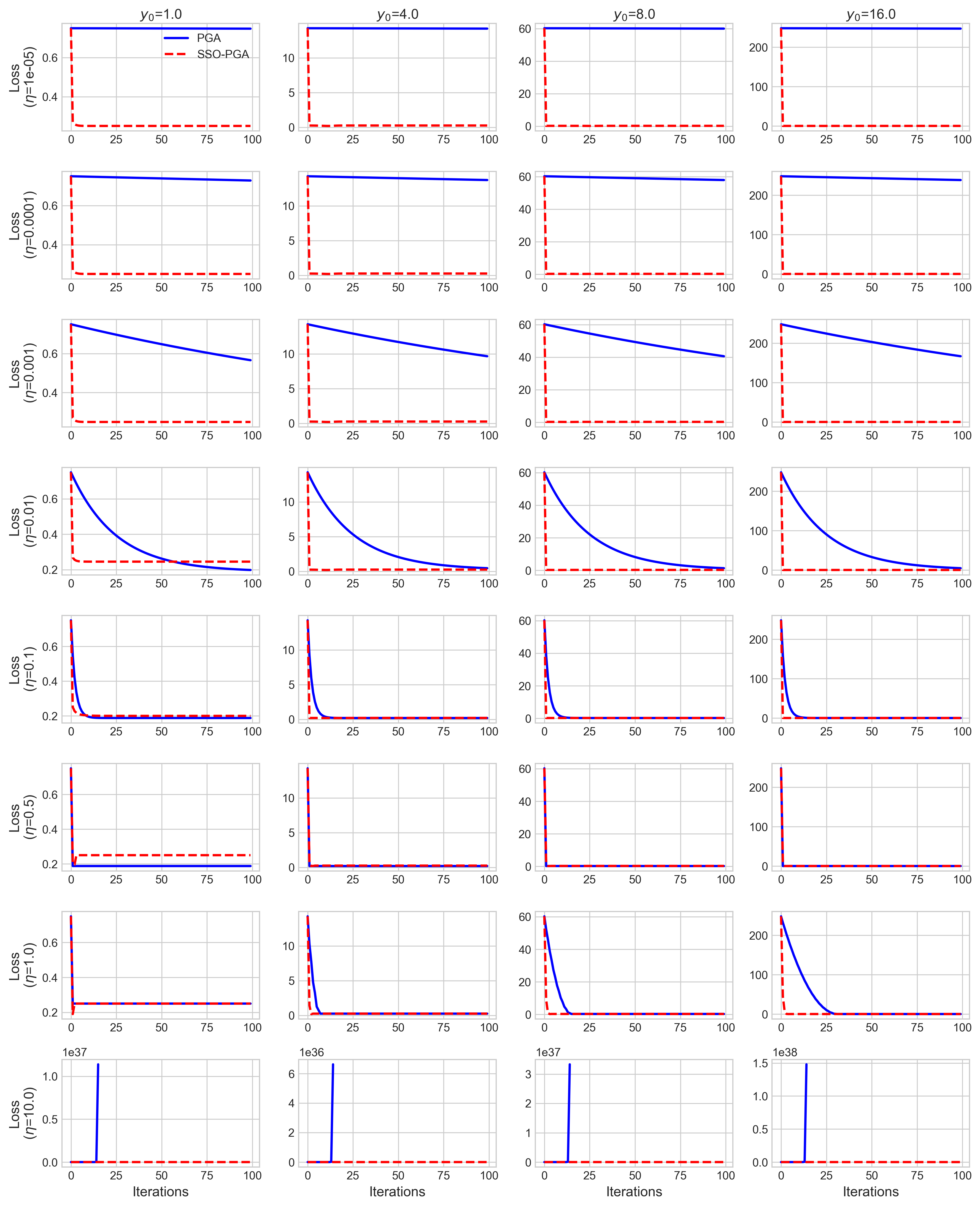}
\caption{Additional comparison of numerical simulation results for SSO-PGA and PGA on Problem II.}
\label{fig:Numerical_Experiments3}
\end{figure}

\begin{figure}[H]
\centering
 \includegraphics[width = \textwidth]{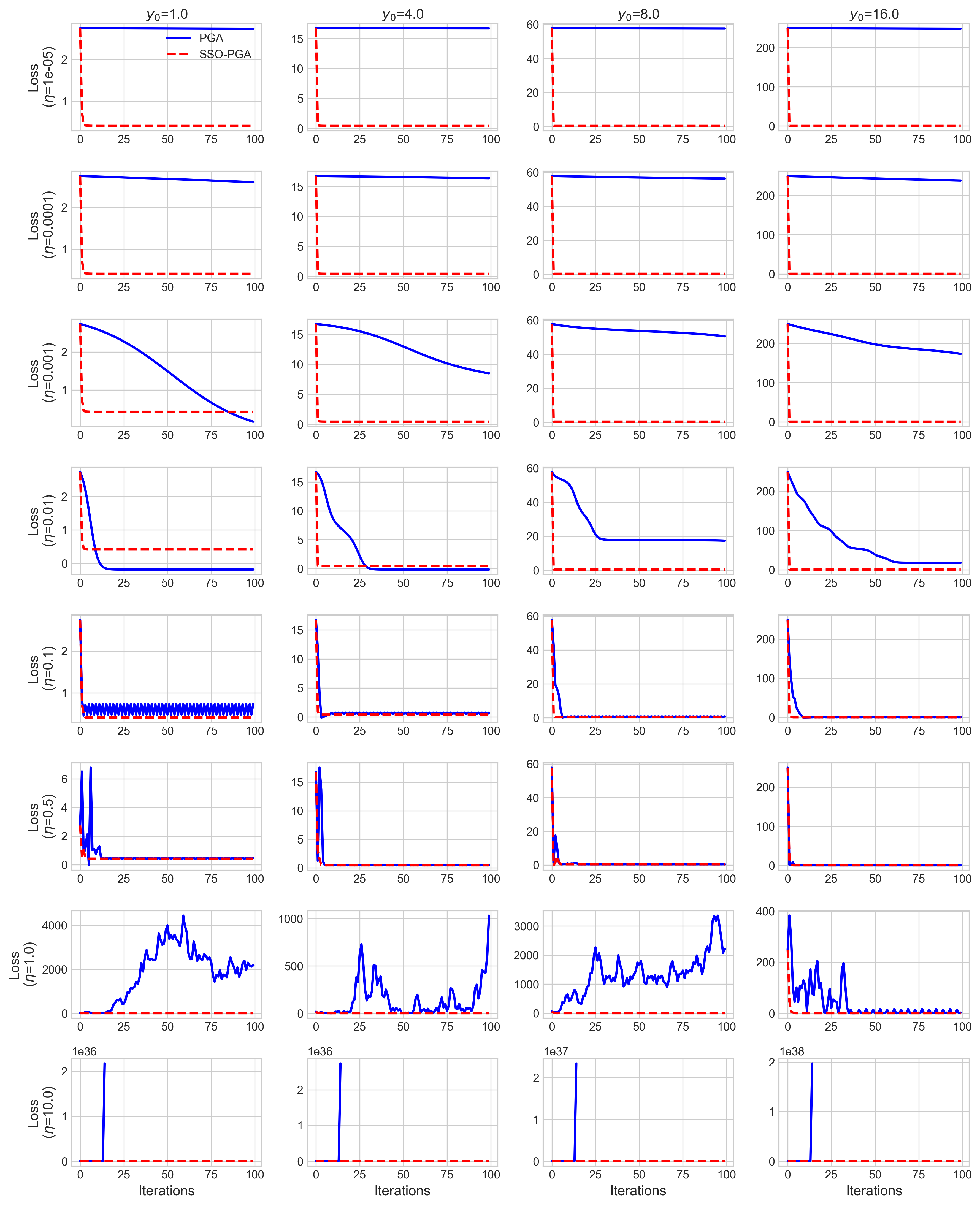}
\caption{Additional comparison of numerical simulation results for SSO-PGA and PGA on Problem II+.}
\label{fig:Numerical_Experiments4}
\end{figure}

Fig.~\ref{fig:function} shows the landscapes for Problem I (left) and Problem I+ (right), respectively. We tested a wide range of learning rates: 1e-5, 1e-4, 1e-3, 1e-2, 1e-1, 5e-1, 1, and 10. As shown in Fig.~\ref{fig:Numerical_Experiments1} to Fig.~\ref{fig:Numerical_Experiments4}, our SSO-PGA consistently outperforms the traditional PGA under most parameter settings. This holds true for both convex and non-convex problems (Problem I and II, and their non-convex counterparts). We can observe that SSO-PGA is less sensitive to the learning rate. When the learning rate is small, SSO-PGA converges much faster than PGA. When the learning rate is large, SSO-PGA is more stable than PGA, especially with very large learning rates where PGA fails to converge. Additionally, in non-convex scenarios, SSO-PGA shows a slight advantage in avoiding local minima.

This success is a direct result of the inherent advantages of the multiplicative update rule introduced by our novel SSO operator. By replacing the traditional subtractive gradient descent step with a sigmoid-based multiplicative update, our algorithm fundamentally transforms the optimization process, making it more stable, less sensitive to hyperparameters, and capable of achieving superior results. It's important to note that since this paper focuses on non-negative inverse problems, the optimal solutions in our numerical simulations are all greater than zero. If the optimal solution were less than zero, it would fall outside the scope of our study, and SSO-PGA would not be able to solve it.

\subsection{Additional Visual Experimental Results}
\label{Additional Experimental Results}
\begin{figure}[H]
\centering
 \includegraphics[width = 0.95\textwidth]{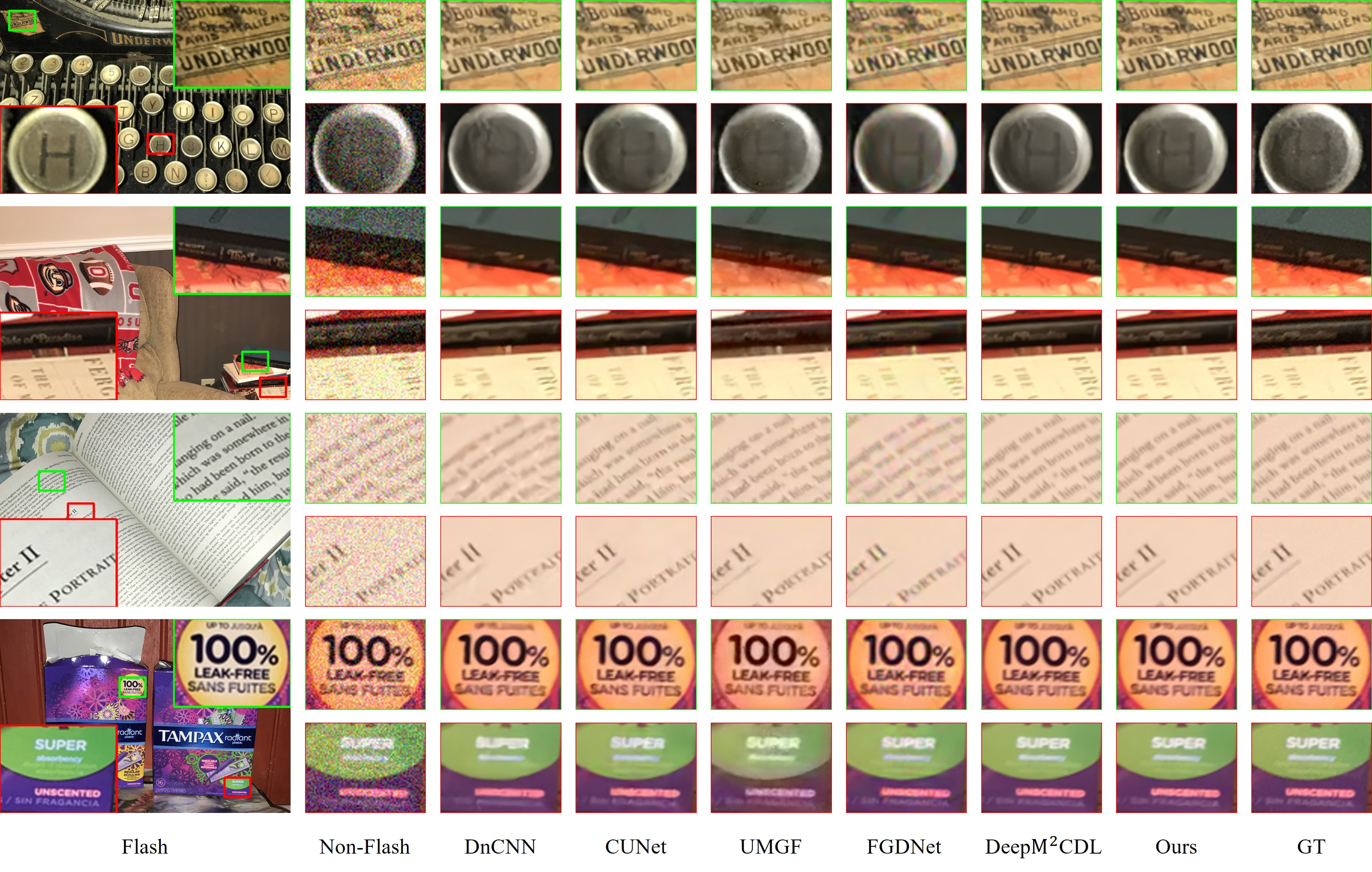}
\caption{Visual comparison of our method and some representative methods on the FAID dataset.}
\label{fig:visual_FAID}
\end{figure}

\begin{figure}[H]
\centering
 \includegraphics[width = 0.95\textwidth]{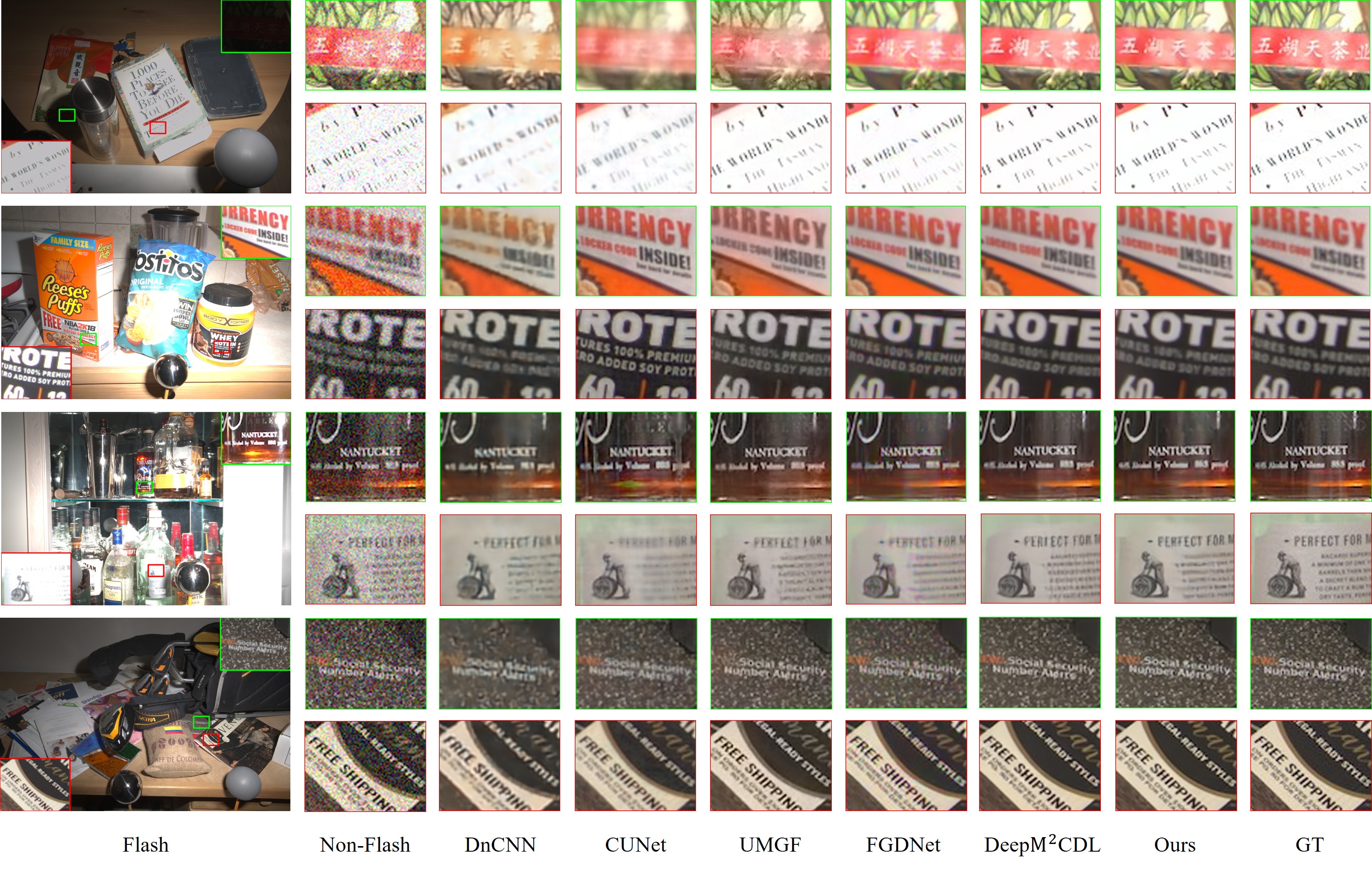}
\caption{Visual comparison of our method and some representative methods on the MID dataset.}
\label{fig:visual_MID}
\end{figure}

\begin{figure}[H]
\centering
 \includegraphics[width = \textwidth]{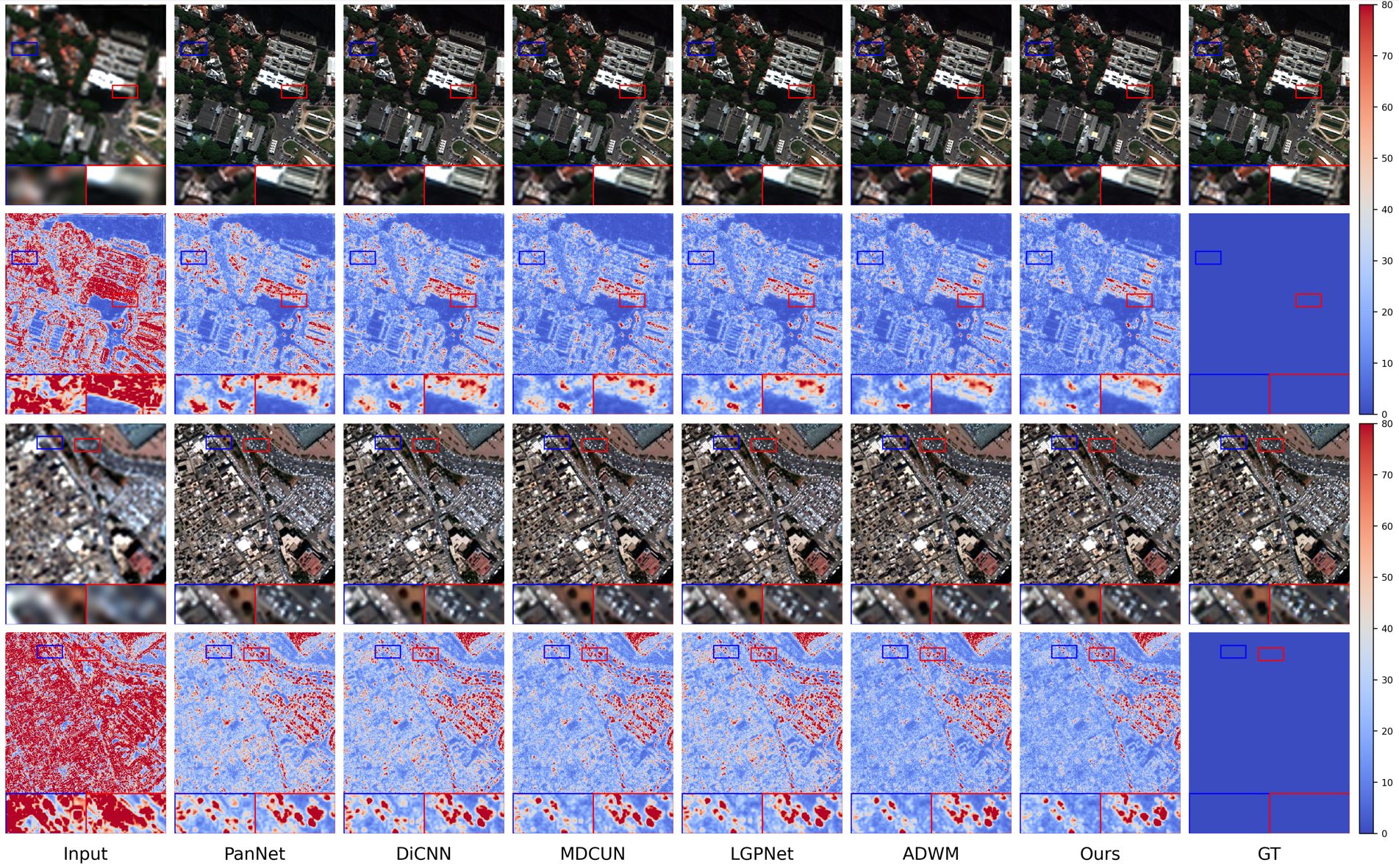}
\caption{Visual comparison (the first row) and the corresponding error map (the second row) of our method and some representative methods on the WV3 reduced-resolution dataset.}
\label{fig:visual_wv3}
\end{figure}

\begin{figure}[H]
\centering
 \includegraphics[width = \textwidth]{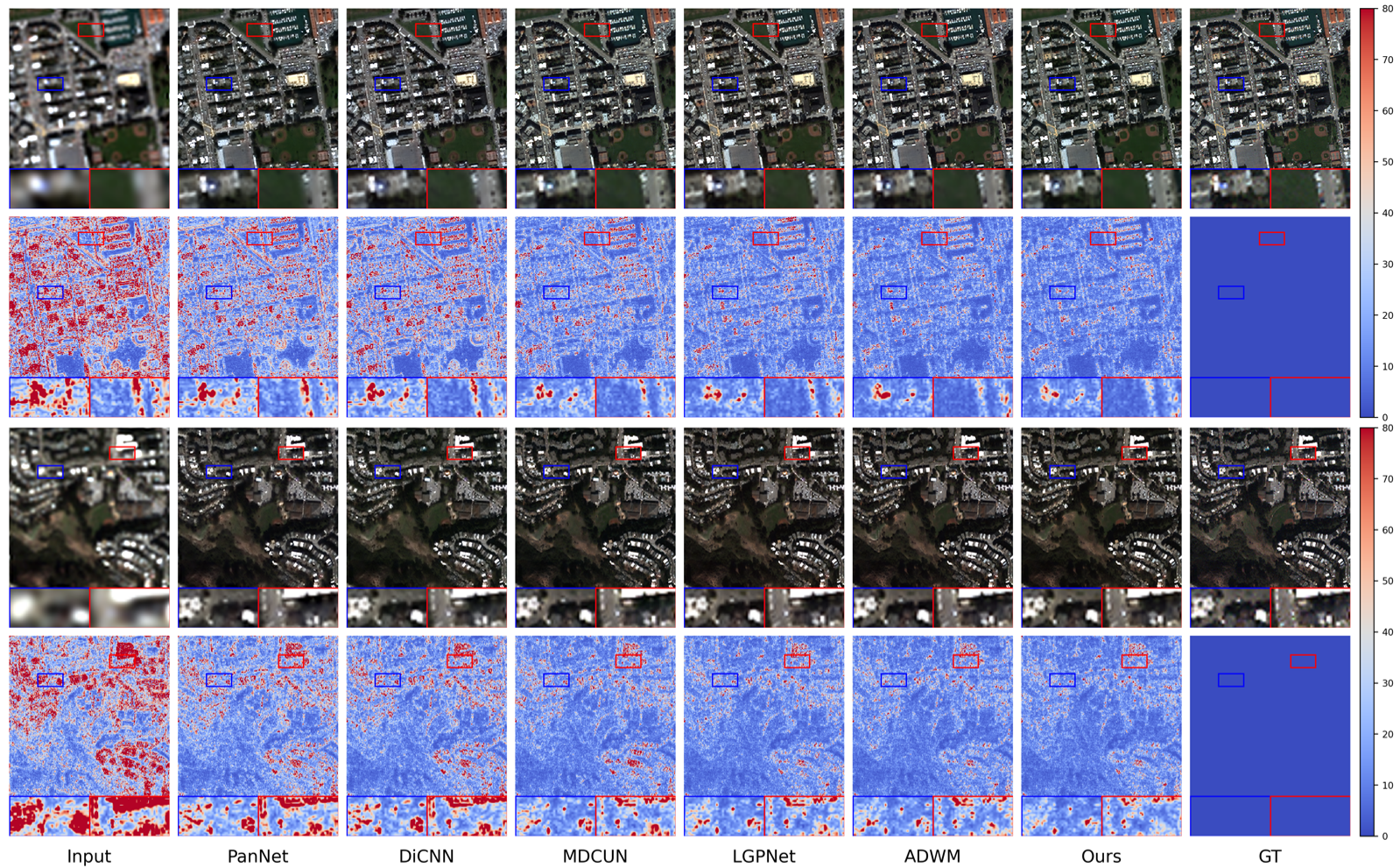}
\caption{Visual comparison (the first row) and the corresponding error map (the second row) of our method and some representative methods on the QB reduced-resolution dataset.}
\label{fig:visual_qb}
\end{figure}

\begin{figure}[H]
\centering
 \includegraphics[width = \textwidth]{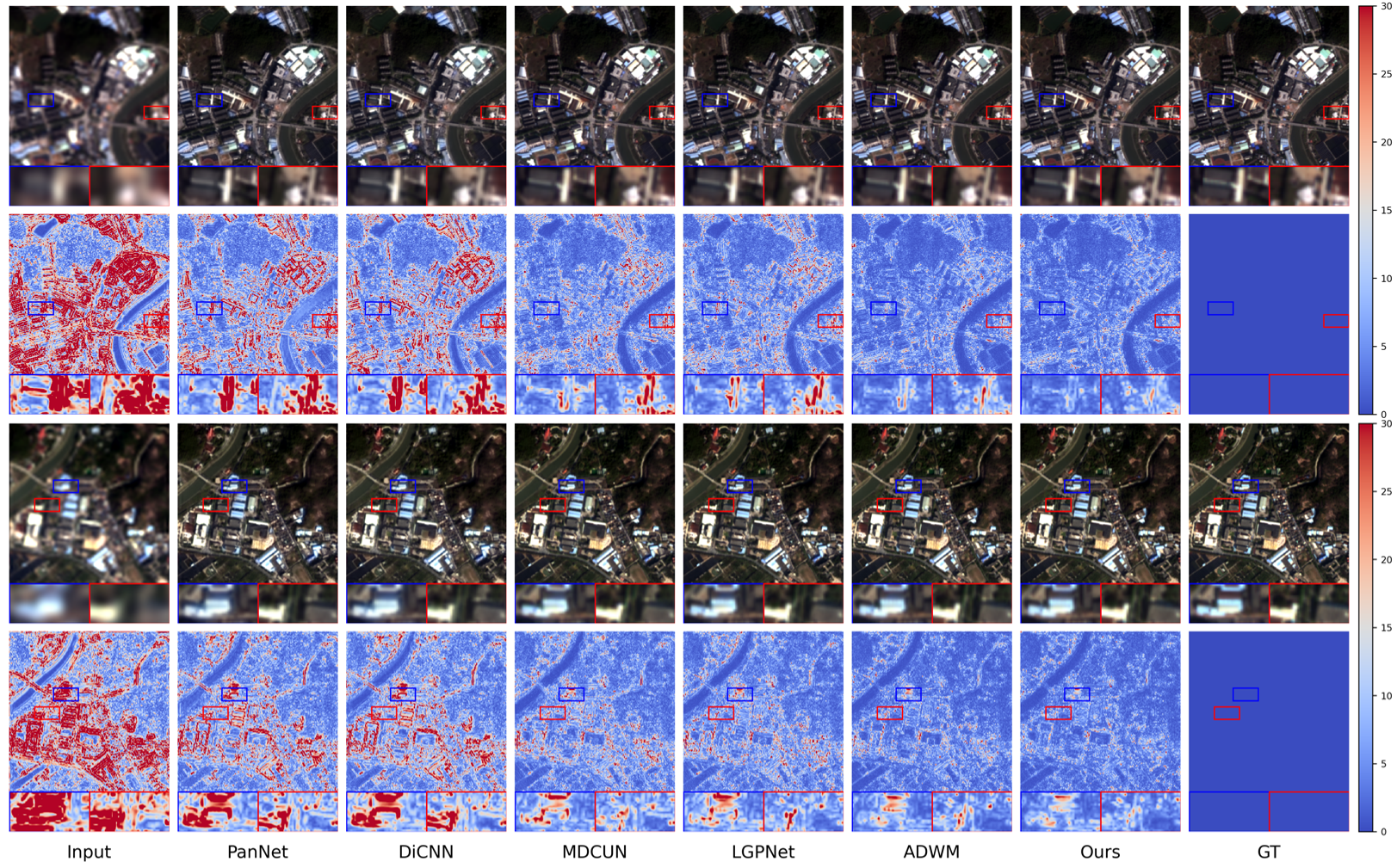}
\caption{Visual comparison (the first row) and the corresponding error map (the second row) of our method and some representative methods on the GF2 reduced-resolution dataset.}
\label{fig:visual_gf}
\end{figure}

\begin{figure}[H]
\centering
 \includegraphics[width = 0.875\textwidth]{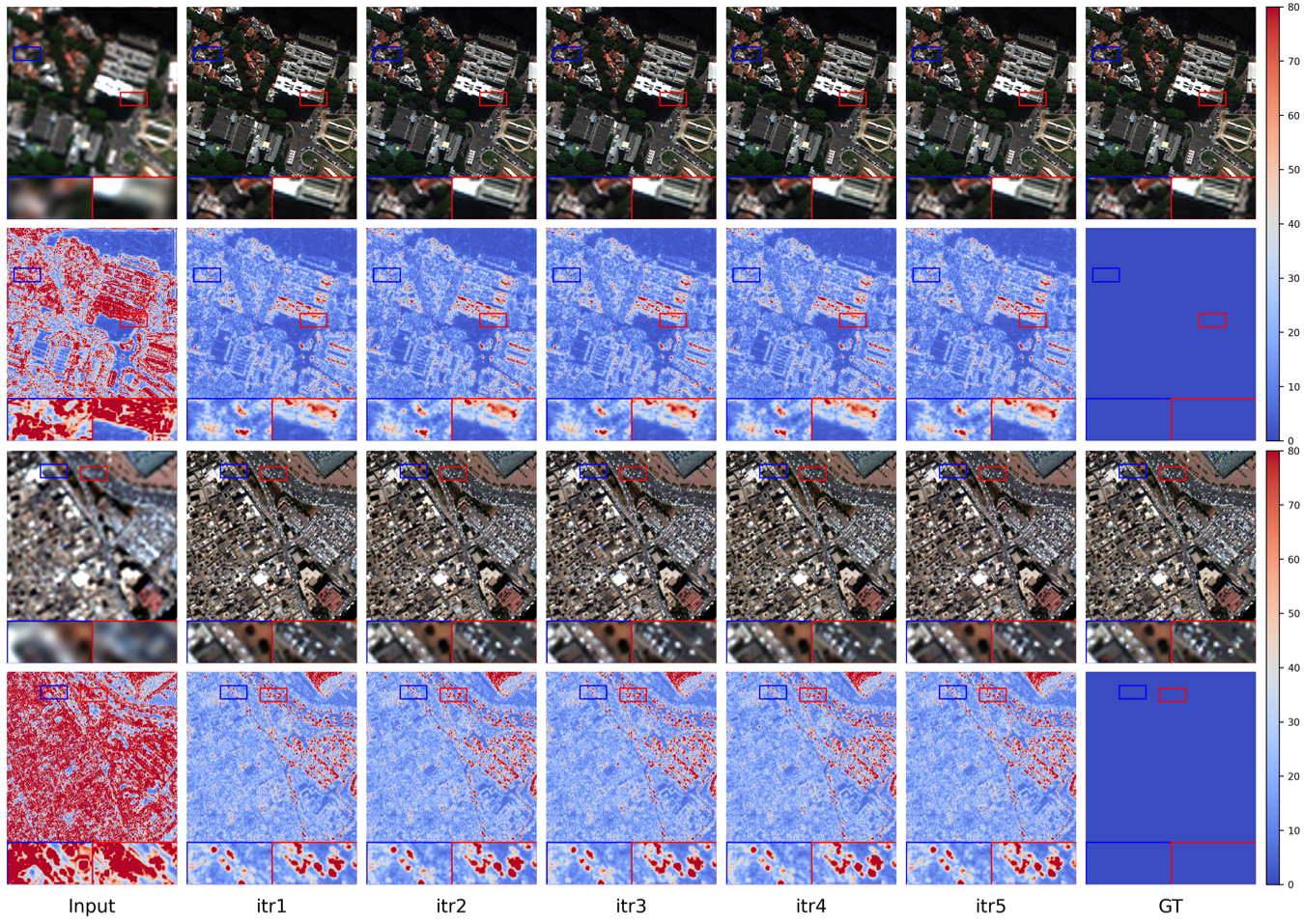}
\caption{Visual comparison (the first row) and the corresponding error map (the second row) of PGA baseline under different iteration steps on the WV3 reduced-resolution dataset.}
\label{fig:itr_old}
\end{figure}

\begin{figure}[H]
\centering
 \includegraphics[width = 0.875\textwidth]{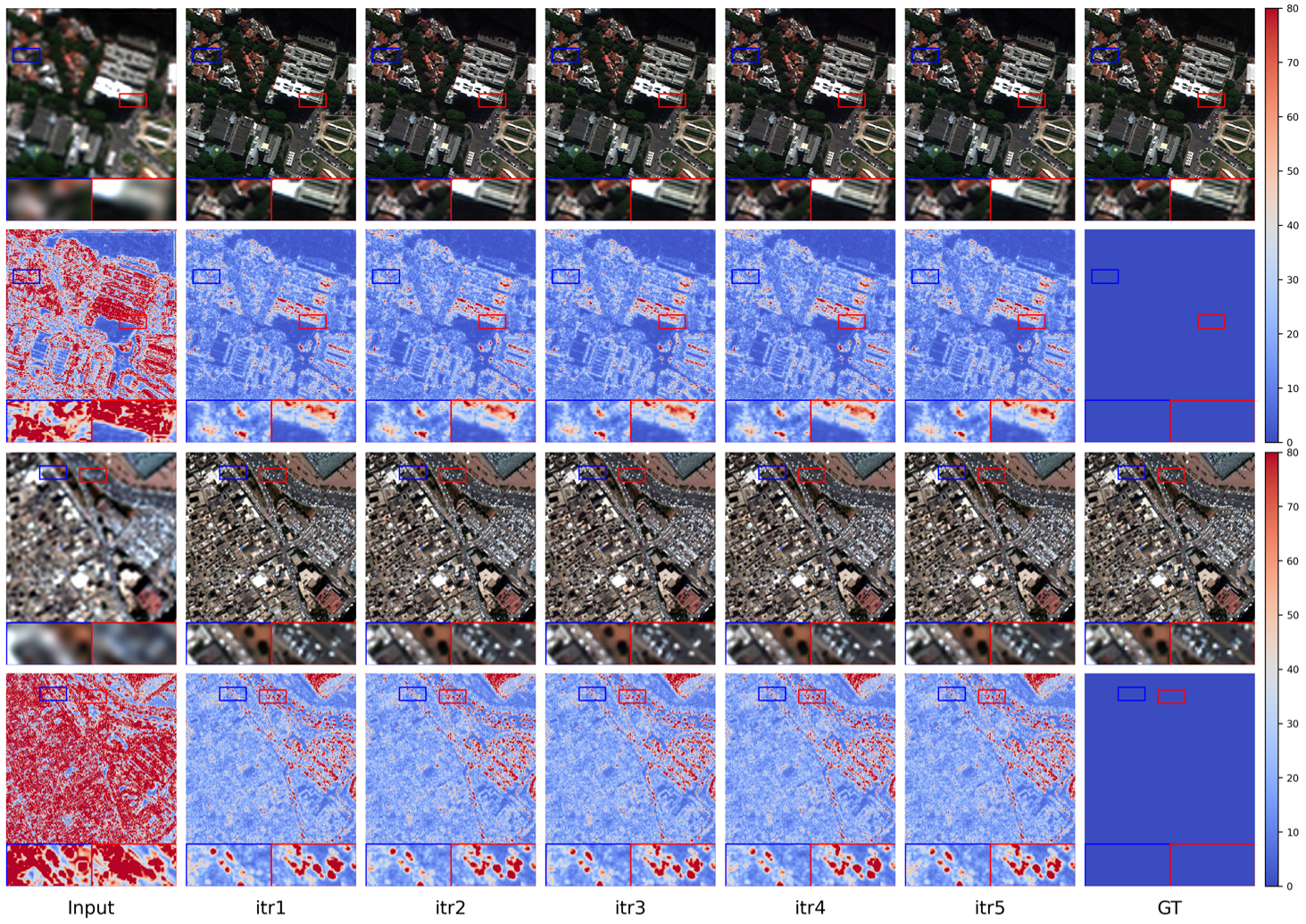}
\caption{Visual comparison (the first row) and the corresponding error map (the second row) of our SSO-PGA under different iteration steps on the WV3 reduced-resolution dataset.}
\label{fig:itr_sso}
\end{figure}

\begin{figure}[H]
\centering
 \includegraphics[width = 0.75\textwidth]{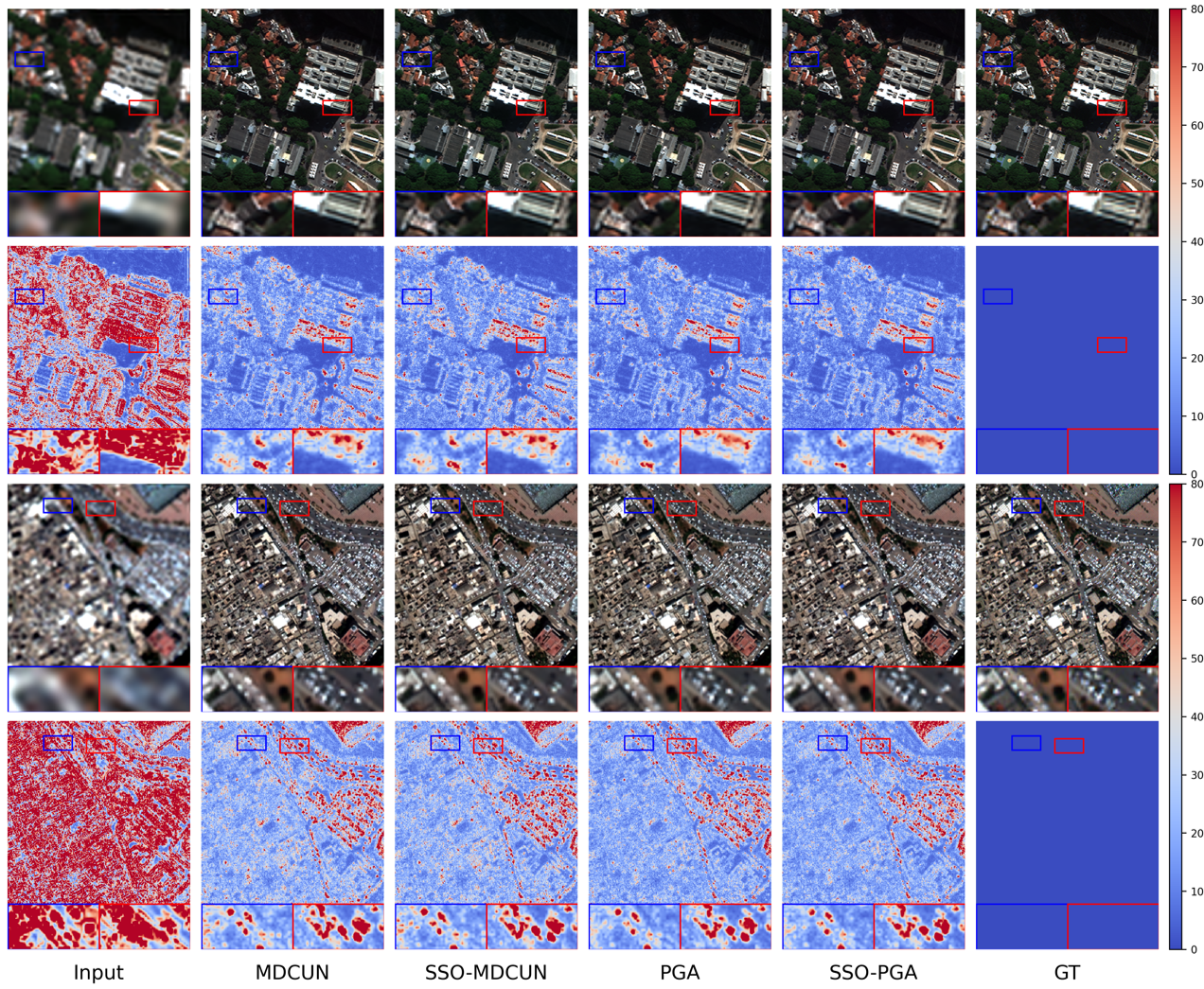}
\caption{Visual comparison (the first row) and the corresponding error map (the second row) of SSO-PGA vs. PGA baseline, and SSO-MDCUN vs. MDCUN on the WV3 reduced-resolution dataset.}
\label{fig:sso_wv3}
\end{figure}

\begin{figure}[H]
\centering
 \includegraphics[width = 0.75\textwidth]{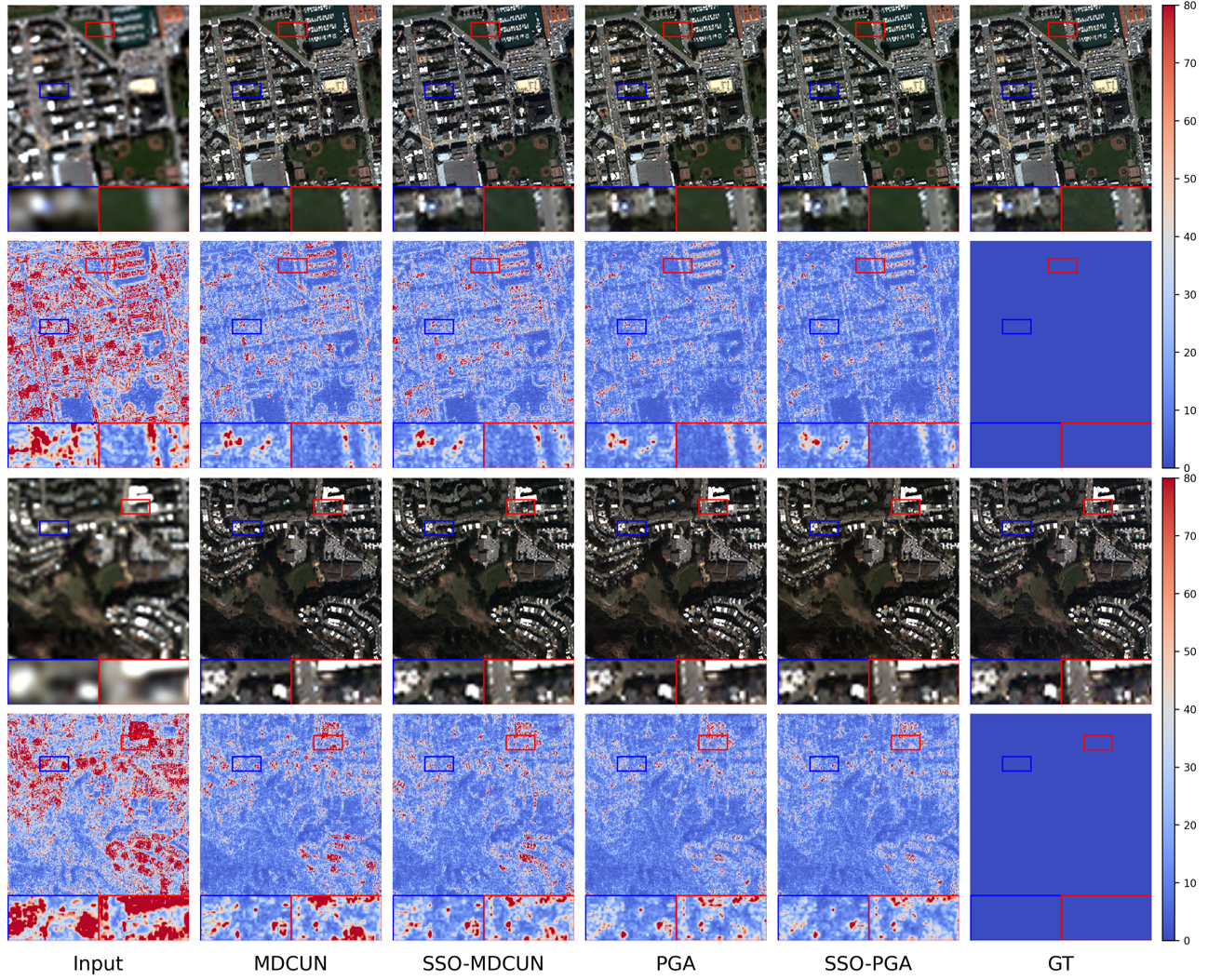}
\caption{Visual comparison (the first row) and the corresponding error map (the second row) of SSO-PGA vs. PGA baseline, and SSO-MDCUN vs. MDCUN on the QB reduced-resolution dataset.}
\label{fig:sso_qb}
\end{figure}

\begin{figure}[H]
\centering
 \includegraphics[width = 0.75\textwidth]{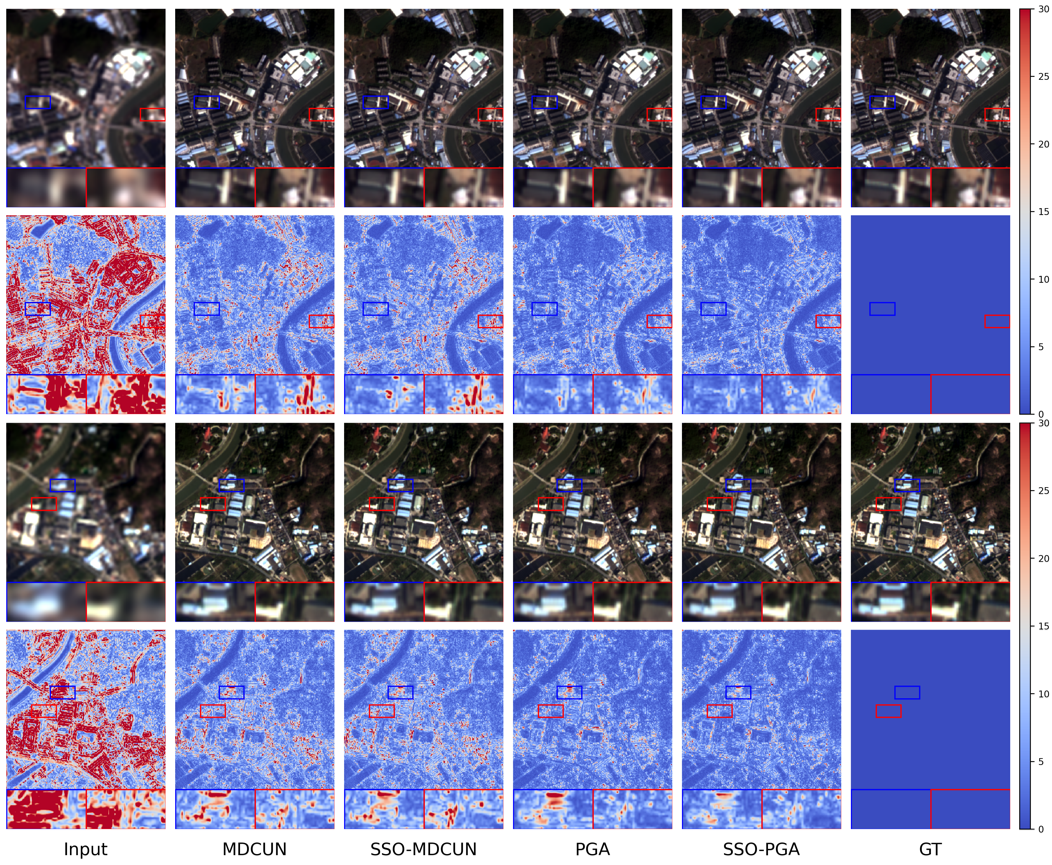}
\caption{Visual comparison (the first row) and the corresponding error map (the second row) of SSO-PGA vs. PGA baseline, and SSO-MDCUN vs. MDCUN on the GF2 reduced-resolution dataset.}
\label{fig:sso_gf}
\end{figure}

In this subsection, we present additional experimental results to further demonstrate the effectiveness and robustness of our proposed SSO-PGA method. The results cover the following aspects:

\begin{itemize}
    \item \textbf{Qualitative Comparison on Flash Guided Non-Flash Image Denoising (Fig.~\ref{fig:visual_FAID}, and Fig.~\ref{fig:visual_MID}:} Visual comparisons between SSO-PGA and several representative SOTA methods are provided across the two benchmark datasets (FAID and MID). These results clearly demonstrate that SSO-PGA consistently achieves superior denoising performance compared to other methods, yielding results that are closer to the ground truth.
    
    \item \textbf{Qualitative Comparison on Multispectral Image Fusion (Fig.~\ref{fig:visual_wv3}, Fig.~\ref{fig:visual_qb}, and Fig.~\ref{fig:visual_gf}):} Visual comparisons between SSO-PGA and several representative SOTA methods are provided across the three benchmark datasets (WV3, QB, and GF2). These results clearly show that SSO-PGA consistently reconstructs sharper spatial details and produces reconstructions closer to the ground truth with lower residual.

    \item \textbf{Visualization Under Different Iteration Steps (Fig.~\ref{fig:itr_old} and Fig.~\ref{fig:itr_sso}):} We further present the reconstructed outputs of both SSO-PGA and the PGA baseline under varying numbers of iterations. The results demonstrate that SSO-PGA achieves high-fidelity fusion even with fewer unfolding steps and maintains performance when increasing the number of iterations, unlike the PGA baseline, which may suffer from degradation.

    \item \textbf{SSO vs. Gradient Descent Visualization (Fig.~\ref{fig:sso_wv3}, Fig.~\ref{fig:sso_qb}, and Fig.~\ref{fig:sso_gf}):} We provide side-by-side visual comparisons of SSO-based and gradient-descent-based models, namely SSO-PGA vs. PGA baseline, and SSO-MDCUN vs. MDCUN \cite{yang2022memory}, across all datasets. The SSO-enhanced variants consistently produce better reconstruction with fewer spectral distortions and residual artifacts.
\end{itemize}

These extended experimental results collectively confirm the superiority of our proposed SSO-PGA framework in terms of reconstruction accuracy, convergence stability, and robustness across different scenarios.

\end{document}